\documentclass[10pt,twocolumn,letterpaper]{article}
\usepackage[T1]{fontenc}
\usepackage[utf8]{inputenc}
\usepackage{times}
\usepackage[letterpaper,top=0.8in,bottom=0.85in,left=0.75in,right=0.75in,columnsep=0.25in]{geometry}
\usepackage[numbers,sort&compress]{natbib}
\usepackage{amsmath,amssymb,graphicx,booktabs,multirow,xspace,pifont}
\usepackage{microtype,caption,cuted,currfile,url,xcolor}
\definecolor{linkblue}{rgb}{0.13,0.32,0.55}
\usepackage[breaklinks,colorlinks,allcolors=linkblue]{hyperref}
\usepackage[capitalize,noabbrev]{cleveref}
\hypersetup{pdftitle={One Fact, Many Places: Redundant Evidence in Long Visual Documents Breaks Single-Gold Evaluation of Retrieval, Citation and Attribution},pdfauthor={},pdfsubject={Research preprint}}
\newcommand{\cmark}{\ding{51}}
\newcommand{\xmark}{\ding{55}}
\newcommand{\eg}{\emph{e.g.}\xspace}
\newcommand{\ie}{\emph{i.e.}\xspace}
\newcommand{\MLB}{MMLongBench-Doc\xspace}
\newcommand{\MDR}{MMDocRAG\xspace}
\newcommand{\EES}{equivalent evidence set\xspace}
\newcommand{\dred}{\ensuremath{\Delta_{\text{red}}}\xspace}
\newcommand{\Mx}[1]{M\textsubscript{#1}}
\newcommand{\maketitlesupplementary}{\clearpage\twocolumn[{\centering\Large\bfseries Supplementary Material\par\vspace{1em}}]}
\title{One Fact, Many Places: Redundant Evidence in Long Visual Documents\\Breaks Single-Gold Evaluation of Retrieval, Citation and Attribution}
\newcommand{\PaperAuthorNames}{Xiaokang Yuan; Binhui Yao; Luwei Yang; Ruimin Chen; Zeng Wang}
\author{%
  Xiaokang Yuan\textsuperscript{1}\quad
  Binhui Yao\textsuperscript{1}\quad
  Luwei Yang\textsuperscript{2}\quad
  Ruimin Chen\textsuperscript{2}\quad
  Zeng Wang\textsuperscript{2}\\[5pt]
  {\small \textsuperscript{1}Zhejiang University, Hangzhou, China}\\
  {\small \textsuperscript{2}Zhejiang Geely Holding Group Co., Ltd, Hangzhou, China}\\[3pt]
  {\small \href{mailto:22621230@zju.edu.cn}{\texttt{22621230@zju.edu.cn}}}%
}
\date{}
\hypersetup{pdfauthor={\PaperAuthorNames}}

\begin{document}
\maketitle
\begin{abstract}
Document understanding is increasingly judged on where an answer comes from, scored against annotated gold locations
as if each fact lived in one place. In long visual documents, a fact often lives in several places: a chart
value is restated in the text, a key figure reappears in the summary. We ask how often such \emph{copies} occur in
standard benchmarks and what they do to evaluation. We annotate 420 questions from \MLB and \MDR with the locations
that independently establish each required fact, checked by two independent reviewers. Copies are common and often
cross-modal: 19.5\% of \MLB test questions have a strict copy outside the gold, and in 28\% of \MDR questions a
non-gold candidate alone supplies a required fact.
Single-gold scoring then fails in three predictable ways. A retriever that finds a copy is scored as a miss: page
recall@1 of text and visual retrievers is underestimated by 13--24 points on questions with a cross-page copy. A model
that cites a copy is scored as citing wrongly: crediting equivalent evidence reverses 11\% of pairwise orderings among
65 competitive \MDR systems.
Masking the gold region leaves the fact available elsewhere: with a copy in the input, 41--56\% of correct answers
across four models survive it yet fail once every annotated instance is masked. Set-level masking with automatically
found copies closes 78\% of the gap to masking every annotated instance at matched granularity (three models); an
unverified key-string rule lowers survival at least as far but masks three times the area. We release the annotations and redundancy-aware
metrics.
\end{abstract}

\section{Introduction}
\label{sec:intro}

\begin{figure*}[t]
  \centering
  \includegraphics[width=\textwidth]{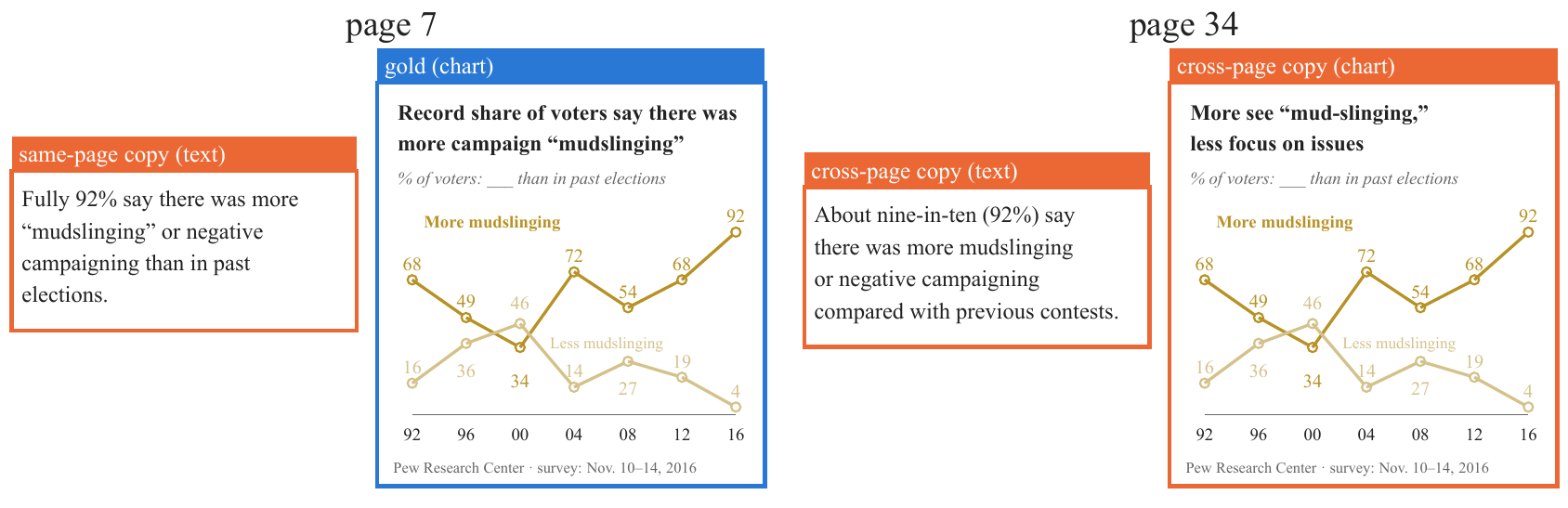}
  \caption{\textbf{One fact, four places.} \MLB asks for the highest share of voters who saw ``more mudslinging than in
  past elections'' (92). The annotated gold is the chart on page 7 (blue). The same value is stated in the text of
  page 7 and again in a chart and a sentence on page 34 (orange). A retriever that returns page 34 is scored as a miss;
  a model that cites page 34 is scored as citing wrongly; and masking the gold chart leaves the answer recoverable,
  so a single-region masking test declares the chart unnecessary although the answer depends on the fact it shows. Source excerpts are redrawn for readability.}
  \label{fig:teaser}
\end{figure*}

Evaluation of document understanding increasingly asks not only \emph{whether} an answer is right but \emph{where} it
comes from. Retrieval for document RAG is scored by whether the gold page is retrieved~\cite{faysse2025colpali,yu2025visrag,cho2024m3docrag};
multimodal RAG benchmarks score which candidate quotes a model cites~\cite{dong2025mmdocrag};
recent attribution benchmarks score the region a model points to and often build the gold region by masking
candidate elements and checking which one the answer depends on~\cite{degrandis2026docattribench,ma2026citevqa};
and selective-answering methods mask the region a model cites and abstain when the answer does not change~\cite{kotte2026evict}.
All of these assume that each required fact lives in \emph{one} place.

Long, visually rich documents often violate this assumption. Reports restate their charts in prose and repeat key
findings in summaries and appendices, and financial filings give the same figure in several statements
(\cref{fig:teaser}). We call the non-gold locations of a fact its \emph{copies}. With copies, single-gold evaluation
makes three kinds of error. A retriever that finds a copy is counted as a miss. A model that cites a copy is counted as
citing wrongly and as missing the gold. Masking the gold location leaves the fact available elsewhere, so the answer
survives and the location is judged unnecessary. That verdict is correct for the region, but it is wrong to read it as
showing that the answer does not use the document.

Redundant evidence has been recognised in text retrieval over high-overlap corpora~\cite{cho2026rare}, in
fact-verification datasets that list alternative evidence sets~\cite{thorne2018fever}, and as a known failure case of
masking-based abstention~\cite{kotte2026evict}. How often it occurs in the human-annotated benchmarks that the
vision--language community uses for long documents, where copies are frequently \emph{cross-modal} (a chart and the
sentence that restates it), and how much it changes citation scores and masking-based attribution, has not been
measured. This paper is such an audit. We also separate what a masking test can establish. Removing one region
tests whether that region is \emph{necessary}, and removing every instance of a fact tests whether the answer
\emph{depends on the fact}. Neither identifies which instance the model read (\cref{sec:formulation}).

\paragraph{Contributions.}
\begin{itemize}
  \item An audit of 220 \MLB~\cite{ma2024mmlongbench} and 200 \MDR~\cite{dong2025mmdocrag} questions in terms of
  \emph{fact units} and their \emph{\EES{}s}, with a protocol that separates strict copies from loose restatements, joint
  support and contextual gold, checked by two independent reviewers. Copies affect roughly one in five to one in three
  questions and concentrate in reports, financial filings and papers (\cref{sec:formulation,sec:measurement}).
  \item A quantification of how single-gold scoring mis-measures retrieval and citation: underestimated recall of text
  and visual retrievers on questions with copies, mis-scored citations that are mostly cross-modal, and changed pairwise
  orderings among 91 published systems; answer-string matching and two open VLM judges do not reproduce the
  equivalent sets in our setting (\cref{sec:retrieval,sec:quotes}).
  \item Evidence, across three VLMs and a text model, that single-region masking misses fact-level dependence for
  41--56\% of correctly answered copy items, that ablation-based attribution splits credit among copies, and that a
  benchmark whose gold is built by masking under-represents
  questions with copies (\cref{sec:masking}).
  \item Set-level masking with automatically detected copies, which recovers most of the effect of the annotated sets
  and is not explained by the extra context it removes; an unverified key-string rule lowers survival as far but masks
  three times the area, and verification raises precision. We release redundancy-aware metrics and the annotations (\cref{sec:method}).
\end{itemize}

\section{Related work}
\label{sec:related}

\paragraph{Long and visually rich document understanding.}
Document VQA moved from single pages~\cite{mathew2021docvqa} to multi-page inputs~\cite{tito2023mpdocvqa,tanaka2023slidevqa}
and to long documents with charts, tables and layout-dependent evidence~\cite{ma2024mmlongbench,deng2025longdocurl}.
These benchmarks annotate evidence pages and sources per question, which later work reuses as retrieval and
attribution gold. We audit such annotations for completeness rather than correctness of the answer.

\paragraph{Retrieval and RAG over document images.}
Page-level retrievers embed page images directly~\cite{faysse2025colpali,yu2025visrag} and are combined with VLM
readers~\cite{cho2024m3docrag,tanaka2025vdocrag}. \MDR~\cite{dong2025mmdocrag} evaluates multimodal RAG by asking models to select and
cite quotes from 15 or 20 retrieved candidates, scoring selection with F1 against gold quotes. Because candidates are
retrieved for relevance, they are exactly where copies of the gold concentrate.

\paragraph{Citation and evidence attribution.}
Text generation with citations is evaluated against retrieved passages~\cite{gao2023alce}, recently with principles that
judge a citation in its full context~\cite{xu2025citeeval}; a citation that supports a statement need not be the one
the model relied on~\cite{wallat2025correctness}. For documents, VISA answers with boxes on page
screenshots~\cite{ma2025visa}, and CiteVQA asks
for element-level box citations whose gold is produced by an automated pipeline that ``identifies crucial evidence via
masking ablation''~\cite{ma2026citevqa}; its overall score also accepts semantically relevant citations, so a valid
non-gold citation need not be penalised. DocAttriBench builds grounding labels on single gold pages by masking layout
elements and measuring the increase in answer perplexity, and filters ambiguous cases, including information repeated
on the page, by score thresholds~\cite{degrandis2026docattribench}.
Training-free attribution methods read attention instead~\cite{tran2026multattnattrib}, or estimate each source's
effect from context ablations~\cite{cohenwang2024contextcite,liu2025attribot,li2026arcjsd}. EVICT masks the region a VLM
cites, re-queries, and abstains when the answer is unchanged; its authors list duplicated evidence as a case in which this
test ``will spuriously signal evidence-independence''~\cite{kotte2026evict}, and report that the test is contraindicated on
chart QA, where answers that survive masking are mostly correct. Masking-based attribution
inherits a known weakness of removal-based explanation: when inputs are redundant, removing one of them changes little,
so its importance is underestimated~\cite{hooker2019roar,deyoung2020eraser}. We quantify this weakness for document
evidence, for single-region masking and for an ablation-based method, and test a set-level mitigation.

\paragraph{Incomplete and alternative gold.}
Incomplete relevance judgments are a long-standing problem in IR evaluation, handled by measures that skip unjudged
documents~\cite{buckley2004bpref,sakai2007bpref}; shallow pools bias the judged set~\cite{buckley2006pooling}, and LLM
assessors that fill it~\cite{upadhyay2024umbrela,otero2026hybrid} can agree with humans overall yet misorder the top
systems~\cite{otero2025limitations};
open-domain QA avoids passage labels by counting any passage that contains the answer string~\cite{karpukhin2020dpr}.
FEVER lists alternative evidence sets for a claim~\cite{thorne2018fever}; HotpotQA annotates supporting facts per
question~\cite{yang2018hotpotqa}; RoMQA targets multi-evidence questions~\cite{zhong2023romqa}. Unlabelled positives are a
known source of error in passage retrieval~\cite{qu2021rocketqa}. Closest to us, RARE builds text retrieval benchmarks
over high-similarity corpora with atomic facts, redundancy-aware gold and a retrieve-then-verify construction of
it~\cite{cho2026rare}; our fact units, equivalent sets and copy detection follow the same idea. What we add is
measurement: we audit existing human-annotated \emph{visual} long-document benchmarks, where copies are often
cross-modal, quantify the effect on citation scoring and masking-based attribution, not only on retrieval, and test
whether answer-string matching and LLM judges can stand in for equivalent sets.

\section{Fact units and equivalent evidence}
\label{sec:formulation}

\paragraph{Fact units.}
For a question $q$ over document $D$ with reference answer $a$, a \emph{fact unit} $u$ is an atomic fact needed to derive
$a$ (\eg ``FY2015 net income $=$ 629{,}551''). A ratio question has two units (numerator and denominator); a lookup has one.
Questions whose answer aggregates over the whole document (``how many line plots are in the report?'') have no
location-specific units and are excluded as not applicable.

\paragraph{Instances and copies.}
An \emph{instance} of $u$ is a region of $D$ (a text block, table cell group, chart, figure or layout element) that by
itself establishes $u$. Benchmarks annotate one or a few instances, the \emph{gold}. Any other instance is a
\emph{copy}. A copy is \emph{strict} if it states the same entity and the same value (``92\%'' and ``fully 92\%'');
it is \emph{loose} if it supports the same conclusion at a different precision or scope (``about a third'' for 33\%, a
direction without the value, a subset of the groups). Regions that support $u$ only jointly are a \emph{combination},
not a copy; regions that are cited as gold but state no required fact (a table caption, a slide title) are
\emph{contextual gold}. A pointer such as ``see page 16'' is never an instance.

\paragraph{Copy forms.}
We label each copy by how it arises:
\emph{restatement} (the fact re-expressed elsewhere, including text that restates a chart or table, which we call
\emph{cross-modal}),
\emph{summary} (abstracts, key-findings pages, selected financial data),
\emph{cross-statement} (the same figure in several financial statements),
\emph{verbatim duplicate} (a repeated section, slide or chart), and
\emph{same element family} (repeated headers and form fields).

\paragraph{Equivalent evidence sets.}
The \EES of $u$ is $E(u)=\{$gold instances$\}\cup\{$strict copies$\}$. A question is \emph{redundant} if some required
unit has $|E(u)|\ge2$ with at least one member outside the gold; it is \emph{fully redundant} if every unit has such a copy.

\paragraph{Redundancy-aware scores.}
Given $E(u)$, three single-gold scores have direct set versions.
(i) \emph{Evidence-aware recall@k}: a ranked list of pages hits if, for every unit, some page in $E(u)$ is in the
top-$k$ (instead of every gold page).
(ii) \emph{Quote selection}: with the set $R$ of quotes that are gold or strict copies, precision is $|C\cap R|/|C|$ for
cited quotes $C$; recall credits a gold quote when it is cited \emph{or} all units it supports are covered by cited
quotes (variant C), or measures unit coverage directly (variant R, which also stops requiring contextual gold).
(iii) \emph{Set-level masking}: to test whether an answer depends on $u$, mask every member of $E(u)$, not one instance.

\paragraph{What masking measures.}
Three questions are easily conflated. \emph{Region necessity}: does the answer change when one region is removed?
\emph{Fact dependence}: does it change when every instance of $u$ is removed? \emph{Source attribution}: which instance
did the model read, and is its citation faithful to that instance? Single-region masking answers the first. When $u$ has
a copy, a region can be sufficient yet unnecessary, so ``unnecessary'' is a correct answer to that question; the error is
to read it as ``the answer does not depend on the document'', or to use it to reject a valid citation or to drop the
region from the gold. Set-level masking answers the second. Neither answers the third: if regions $A$ and $B$ both state
$u$, masking $A$, $B$ or both cannot reveal which one the model used. Our masking experiments compare the first two.

\section{Prevalence of copies}
\label{sec:measurement}

\paragraph{Data and protocol.}
\MLB~\cite{ma2024mmlongbench} pairs 1{,}082 questions with 135 long PDFs (median 28 pages over our annotated questions) and annotates
evidence pages and evidence sources. We annotated a 60-question development sample, used to fix the protocol, and an
independent 160-question test sample, stratified by evidence source (text, table, chart, figure, layout, two-page) with at
most two questions per document. For every applicable question we identified the fact units, verified the gold pages, and
searched the whole document for instances: string and number search over the text layer (OCR for 13 scanned slide decks),
reading of candidate contexts, and rendering pages when needed.
\MDR~\cite{dong2025mmdocrag} provides 15 candidate quotes (text passages and images) per question; we annotated 200 random
questions from its evaluation set, recording for every candidate which units it supports alone, loosely, or only jointly.
All annotations were made by one annotator, an LLM (Claude) following a written protocol, so the human review below
is the main check on them. A second search pass over 30 random test questions labelled ``no copy''
(BM25 over the fact context plus number-format variants) found no missed copy.

\paragraph{Independent review.}
Two human reviewers re-annotated 60 questions independently (15 labelled redundant and 15 copy-free per benchmark), blind
to our labels. They agreed with each other at Cohen's $\kappa=0.93$ on both benchmarks and confirmed all 15 of our \MLB
positives and 13 of 15 \MDR positives (the other two are loose copies); at the relation level our strict relations reach
a precision of 0.97--1.00 (\MLB) and 0.74--0.79 (\MDR) against their judgments. All values meet the acceptance
thresholds we fixed before the review (\cref{sec:s-annot}). In the other direction, both reviewers found copies in 2 of
the 15 questions per benchmark that we had labelled copy-free, so errors run mostly toward under-counting: correcting
the rates with the review gives 30--33\% for \MLB and 34--36\% for \MDR. The miss rate rests on 15 questions per
benchmark (95\% CI 4--38\%), so all tables report our original annotation, frozen before the review, and we treat the
corrected rates as indicative.

\begin{table}[t]
  \centering
  \small
  \setlength{\tabcolsep}{3pt}
  \begin{tabular}{@{}lccc@{}}
    \toprule
    Sample & $n$ & Redundant & 95\% CI \\
    \midrule
    \multicolumn{4}{@{}l}{\textit{\MLB: a strict copy outside the gold}} \\
    Test (160, stratified) & 154 & 19.5\% & 14.0--26.4 \\
    \quad reweighted to the pool & -- & 20.4\% & 14.2--27.2 \\
    \quad every unit has a copy & 154 & 14.9\% & 10.2--21.4 \\
    \quad incl.\ loose copies & 154 & 22.7\% & 16.8--30.0 \\
    Development (60) & 54 & 42.6\% & 30.3--55.8 \\
    \midrule
    \multicolumn{4}{@{}l}{\textit{\MDR (200 questions)}} \\
    Non-gold quote suffices (strict) & 196 & 28.1\% & 22.2--34.7 \\
    \quad incl.\ loose & 196 & 41.8\% & 35.2--48.8 \\
    A unit has $\ge2$ supporting quotes & 196 & 52.0\% & 45.1--58.9 \\
    Gold quotes copy each other & 196 & 45.9\% & 39.1--52.9 \\
    Contextual gold present & 196 & 33.7\% & 27.4--40.5 \\
    \bottomrule
  \end{tabular}
  \caption{\textbf{Prevalence of copies.} Wilson intervals. The development sample was annotated first and is
  reported separately; the test sample is the primary estimate.}
  \label{tab:prevalence}
\end{table}

\begin{figure}[t]
  \centering
  \includegraphics[width=\linewidth]{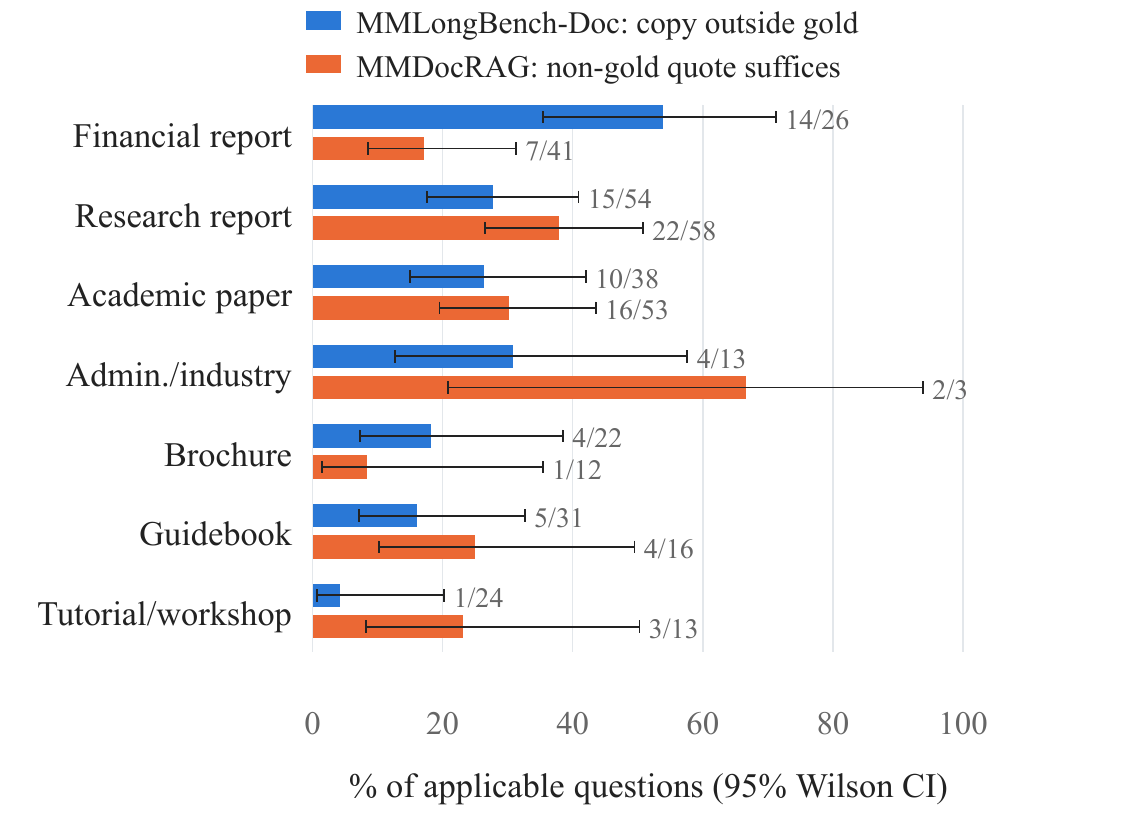}
  \caption{\textbf{Copies by document type} (both samples pooled for \MLB). Bars show the share of applicable questions;
  labels give counts.}
  \label{fig:prevalence}
\end{figure}

\paragraph{Prevalence.}
\Cref{tab:prevalence} gives the main estimates. On the independent \MLB test sample, one in five applicable questions has a
strict copy of a required fact outside the gold; reweighting the strata to the benchmark's composition gives the same
20\%. The development sample shows a higher rate (43\%), a gap larger than sampling noise (Fisher $p=0.002$). It was
annotated first, while the protocol was being fixed, and later re-checked under the final strict rule (two questions
were downgraded to loose); we cannot exclude that it remains more inclusive, or that its documents differ by chance. We
therefore treat the test sample as the estimate and the development rate as an upper reference. Analyses that pool both
samples (\cref{fig:prevalence}, \cref{sec:retrieval}) say so; their test-only values are in the supplement (\cref{sec:s-test}). In \MDR, where candidates are retrieved for relevance, 28\% of questions have a non-gold candidate that
alone supplies a required fact, and in 52\% some fact has more than one supporting candidate when gold quotes that repeat
each other are counted.

\paragraph{Where copies occur.}
Copies concentrate in document types that summarise and restate (\cref{fig:prevalence}): in the pooled \MLB sample,
14 of 26 financial-report questions (54\%) have one (cross-statement and selected-financial-data copies), against 4\%
for tutorial slides. Questions whose
evidence is a single figure rarely have copies (4\%); questions with text, tables or charts do (18--24\%), as do two-page
questions (29\%). Restatement is the most common form, followed by summaries and cross-statement copies; 16 of the 30
redundant test questions have a cross-modal copy, typically a sentence restating a chart value.

\paragraph{Gold annotations are also incomplete.}
In 12 of 220 \MLB questions (5.5\%) the gold pages are wrong: printed page numbers used as PDF indices, off-by-one
offsets, pages that do not contain the fact, and one document that does not match its questions. In 9 of 196 \MDR
questions (4.6\%) a required fact is supported only by non-gold candidates.

\section{What copies do to evaluation}
\label{sec:consequences}

We now score existing systems twice, once against the benchmark's single gold and once against the equivalent sets of
\cref{sec:formulation}. \Cref{tab:main} gives the main results: page retrieval and masking on \MLB, and quote citation on
\MDR.

\begin{table*}[t]
  \centering\footnotesize
  \begin{minipage}[t]{0.475\textwidth}
    \centering
    \textbf{\MLB}\\[2pt]
    \textit{(a) Page retrieval, R@1 (\%), 46 questions with a cross-page copy}\\[1pt]
    \setlength{\tabcolsep}{5pt}
    \begin{tabular}{@{}lcccc@{}}
      \toprule
      Retriever & Gold & Answer & $E(u)$ (ours) & $\Delta$ \\
      \midrule
      BM25 & 19.6 & 34.8 & \textbf{43.5} & +23.9 \\
      E5-base & 19.6 & 30.4 & \textbf{39.1} & +19.6 \\
      ColPali & 37.0 & 39.1 & \textbf{54.3} & +17.4 \\
      ColQwen2 & 43.5 & 39.1 & \textbf{56.5} & +13.0 \\
      \bottomrule
    \end{tabular}\\[6pt]
    \textit{(b) Masking, answer survival (\%), six-page input}\\[1pt]
    \setlength{\tabcolsep}{4.5pt}
    \begin{tabular}{@{}lccccc@{}}
      \toprule
      & \multicolumn{4}{c}{Copy items} & Controls \\
      \cmidrule(lr){2-5}\cmidrule(l){6-6}
      Model & $n$ & M\textsubscript{gold} & M\textsubscript{all} & \dred & M\textsubscript{gold} \\
      \midrule
      Qwen2.5-VL-7B & 29 & 86.2 & 31.0 & 55.2 & 15.7 \\
      InternVL3-8B & 26 & 76.9 & 30.8 & 46.2 & 20.4 \\
      GPT-6.1 Sol & 32 & 93.8 & 53.1 & 40.6 & 33.3 \\
      Qwen3-4B (text) & 25 & 76.0 & 20.0 & 56.0 & 11.4 \\
      \bottomrule
    \end{tabular}
  \end{minipage}\hfill
  \begin{minipage}[t]{0.505\textwidth}
    \centering
    \textbf{\MDR}\\[2pt]
    \textit{(c) Quote citation, Quote F1 and rank among 91 systems}\\[1pt]
    \setlength{\tabcolsep}{3.5pt}
    \begin{tabular}{@{}llcccccc@{}}
      \toprule
      & & \multicolumn{2}{c}{Original} & \multicolumn{2}{c}{C (ours)} & \\
      \cmidrule(lr){3-4}\cmidrule(lr){5-6}
      System & Input & F1 & Rank & F1 & Rank & Change \\
      \midrule
      GPT-4.1 & MM & .736 & 1 & .790 & 1 & -- \\
      GPT-4.1 & Text & .682 & 2 & .758 & 5 & $\downarrow$3 \\
      Qwen2.5-32B-FT & Text & .667 & 3 & .775 & 3 & -- \\
      Gemini-2.5-Pro & MM & .665 & 4 & .718 & 6 & $\downarrow$2 \\
      Qwen2.5-72B-FT & Text & .663 & 5 & .785 & 2 & $\uparrow$3 \\
      Gemini-2.0-Flash-Think & MM & .653 & 6 & .711 & 8 & $\downarrow$2 \\
      GPT-4o & MM & .651 & 7 & .718 & 7 & -- \\
      Grok-3-mini & Text & .649 & 8 & .705 & 10 & $\downarrow$2 \\
      Claude-3.5-Sonnet & MM & .646 & 9 & .697 & 19 & $\downarrow$10 \\
      GPT-4.1-mini & MM & .645 & 10 & .695 & 21 & $\downarrow$11 \\
      Qwen2.5-14B-FT & Text & .629 & 20 & .765 & 4 & $\uparrow$16 \\
      Qwen2.5-32B & Text & .608 & 30 & .708 & 9 & $\uparrow$21 \\
      \midrule
      \multicolumn{2}{@{}l}{Mean of 91 systems} & .533 & & .601 & & \\
      \bottomrule
    \end{tabular}
  \end{minipage}
  \caption{\textbf{Main results: single-gold vs.\ equivalent-set scoring.}
  (a) Gold: every fact unit has a verified gold page at rank 1. Answer: the top page contains the answer string.
  $E(u)$: any page of the equivalent set counts. $\Delta$: $E(u)$ minus Gold. BM25 and E5 read the text layer;
  ColPali and ColQwen2 embed page images.
  (b) Items answered correctly without a mask, masked at their gold instances (M\textsubscript{gold}) or at all
  annotated instances (M\textsubscript{all}). Controls: facts whose key string occurs only in the gold region.
  (c) The 12 systems in the top ten under either metric; all 91 are in \cref{tab:s-systems}. MM: images given as images;
  Text: text-only input. FT: fine-tuned by the \MDR authors. Every $\Delta$ and \dred{} has a 95\% bootstrap interval
  that excludes zero (\cref{tab:s-retr-all,tab:s-masking-full}).}
  \label{tab:main}
\end{table*}

\subsection{Page retrieval}
\label{sec:retrieval}
We rank the pages of each document with two text retrievers, BM25 (text layer with OCR fallback) and
E5-base~\cite{wang2022e5}, and two visual retrievers that embed page images, ColPali and
ColQwen2~\cite{faysse2025colpali}. We use the 208 applicable \MLB questions of both samples. Under single-gold scoring, a
ranking hits at $k$ when every fact unit has a verified gold page in the top $k$; under evidence-aware scoring, any page
of $E(u)$ counts. We also report the answer-string hit used in open-domain QA~\cite{karpukhin2020dpr}, which needs no
evidence annotation.

\Cref{tab:main}(a) shows R@1 on the 46 questions with a strict cross-page copy. Single-gold scoring underestimates all
four retrievers, by 20--24 points for the text retrievers and by 13--17 points for the visual ones, which find the gold
page more often. Over all 208 questions the gap is 3--5 points, and errors in the dataset's own page labels cost another
5--10 points (\cref{tab:s-retr-all}). The answer-string hit does not close the gap. The answer string occurs on no
evidence page for 46\% of the questions, and on the cross-page subset it stays below the evidence-aware score for every
retriever. The test sample alone shows the same pattern (\cref{sec:s-test}).

\subsection{Quote selection in multimodal RAG}
\label{sec:quotes}
\MDR scores the quotes a model cites by F1 against the gold quotes. We re-scored the 91 published system outputs
(15-candidate setting) on our 200 annotated questions, on which the original ranking agrees with the full-benchmark
ranking (Kendall $\tau=0.89$). Our main correction is variant C (\cref{sec:formulation}), which credits a missed gold
quote when cited copies cover its facts. Applying all corrections from the human review changes Quote F1 under C by at
most 0.003 (\cref{sec:s-quotes}).

\paragraph{Per-system results.}
\Cref{tab:main}(c) lists the 12 systems that rank in the top ten under either metric. GPT-4.1 with image input stays
first, but the order below it changes. Claude-3.5-Sonnet drops from 9th to 19th and GPT-4.1-mini from 10th to 21st,
while Qwen2.5-14B-FT, fine-tuned by the \MDR authors, rises from 20th to 4th. The systems that drop cite 4.7 and 5.1
quotes per question on average; the fine-tuned Qwen2.5 models cite 1.5--2.0 (\cref{fig:scatter}). The original metric
rewards citing every copy of a fact and penalises a single sufficient citation. Under C, both receive the same recall.
Among the 65 systems with original F1 $\ge0.5$, C reverses 11.4\% of the pairwise orderings.

\begin{figure}[t]
  \centering
  \includegraphics[width=\linewidth]{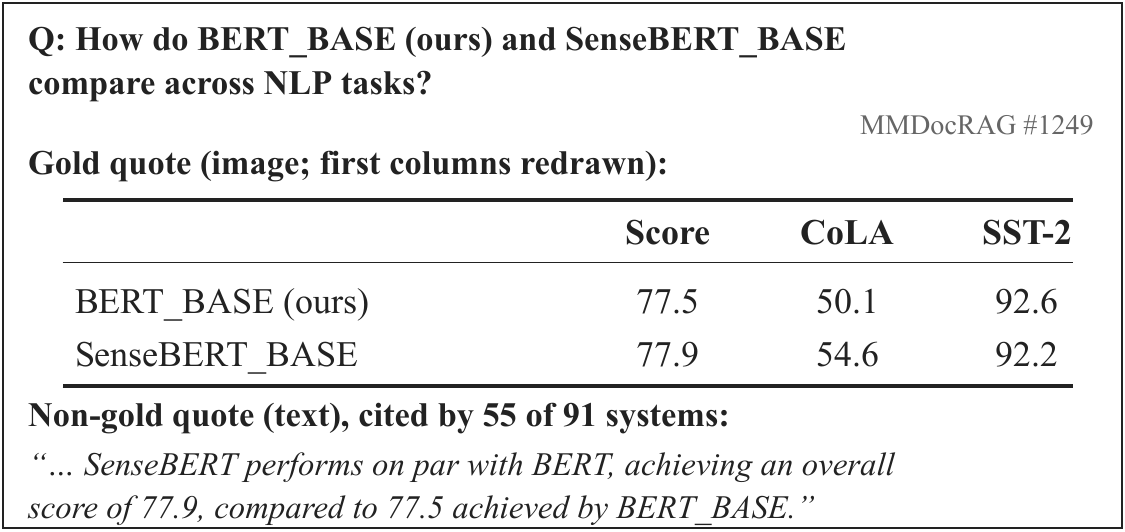}
  \caption{\textbf{A cross-modal copy.} The text passage states the overall scores of the gold table, yet the original
  metric counts every citation of it as wrong. For GPT-4.1, which cites both quotes, Quote F1 on this question is 0.67
  under the original metric and 1.00 under C.}
  \label{fig:example}
\end{figure}

\paragraph{Mis-scored citations.}
Of the citations the original metric counts as false positives, 11.4\% are strict copies of a gold fact (19.1\% with
loose copies). Of the gold quotes it counts as missed, 29.3\% are covered by a cited copy, and another 31.0\% are
contextual gold that states no required fact. Most copies are cross-modal (\cref{fig:example}). In at least 63\% of the
valid non-gold citations, an image restates a text gold quote or the reverse. Non-gold quotes cited by at least half of
the systems are valid evidence in 30\% of cases, against 0.4\% for quotes cited by fewer than a quarter
(\cref{fig:consensus}).

\begin{table}[t]
  \centering
  \small
  \setlength{\tabcolsep}{2pt}
  \begin{tabular}{@{}lccccc@{}}
    \toprule
    Metric & F1 & $\tau$ & $\tau_{\text{comp}}$ & Flips & P\,/\,R \\
    \midrule
    Original (single gold) & .533 & -- & -- & -- & -- \\
    A: credit valid citations & .552 & .966 & -- & -- & -- \\
    C: equivalent-set credit & .601 & .878 & .771 & 11.4\% & .74\,/\,.87 \\
    R: unit recall & .627 & .818 & .670 & 16.5\% & -- \\
    \midrule
    \multicolumn{6}{@{}l}{\textit{C, support relations without our annotation}} \\
    \quad Key-string match & .623 & .933 & .887 & 5.7\% & .21\,/\,.68 \\
    \quad Qwen2.5-VL-7B judge & .566 & .969 & .942 & 2.9\% & .78\,/\,.58 \\
    \quad InternVL3-8B judge & .609 & .908 & .828 & 8.6\% & .65\,/\,.87 \\
    \bottomrule
  \end{tabular}
  \caption{\textbf{Scoring rules for \MDR Quote F1} (mean over 91 systems). A counts valid non-gold citations in
  precision; R also scores recall over fact units. $\tau$: Kendall agreement with the original ranking;
  $\tau_{\text{comp}}$ and flips: among the 65 systems with original F1 $\ge0.5$. The top system is GPT-4.1 under every
  row except R (Qwen2.5-72B-FT). P\,/\,R: precision and recall of the (quote, fact unit) support relations against the
  human review. Bottom: C with the support relations built by string matching or by an open VLM judge.}
  \label{tab:quotes}
\end{table}

\paragraph{Scoring rules.}
\Cref{tab:quotes} separates the parts of the correction. Counting valid non-gold citations in precision (A) hardly
changes the ranking ($\tau=0.97$); most of the change comes from the recall credit in C. The C flip rate is stable under
bootstrap resampling of questions (8.5--14.8\%), 59\% of the flipped pairs differed by at least 0.01 F1 under the
original metric, and the rate rises to 18.4\% among the 34 systems above 0.6. Global rank correlation can hide such
changes~\cite{otero2025limitations}. Variant R changes the ranking more ($\tau_{\text{comp}}=0.67$), but it also drops
the requirement to cite contextual gold. We therefore use C as the redundancy correction and R as an upper bound.

\paragraph{String matching and VLM judges.}
We also built the support relations without our annotation, keeping the rest of the metric fixed (\cref{tab:quotes},
bottom). One baseline matches the numbers and quoted phrases of each fact unit. The other two ask an open VLM,
Qwen2.5-VL-7B or InternVL3-8B, whether a candidate by itself establishes the unit (\cref{sec:s-baselines}). Against the
human review, string matching has low precision (0.21). The judges reach 0.78/0.58 and 0.65/0.87 in precision and
recall, against 0.74/0.87 for our annotation, and neither meets the criterion we set before running them. They agree
with each other at $\kappa=0.67$ and miss 24 and 13 of the 49 cross-modal copies in our annotation. With their
relations, C reverses 2.9\% and 8.6\% of competitive pairs; a different flip rate by itself does not show which ranking
is closer to human judgment. In our setting, these judges do not replace the annotation.

\subsection{Masking-based attribution}
\label{sec:masking}
Attribution benchmarks and selective-answering methods mask a candidate region and check whether the answer
changes~\cite{degrandis2026docattribench,ma2026citevqa,kotte2026evict}. If the fact has a copy in the input, the answer
can survive the mask.

\paragraph{Setup.}
We use a separate set of 157 \MLB fact units whose instances can be located in the text layer, or by a drawn box for
charts without one. Of these, 59 have copies (56 cross-page) and 98 are controls whose key string occurs only
in the gold region. It is a selected set, not a random sample: of its 154 six-page inputs, 81 come from our test
sample, 19 from the development sample and 54 from other benchmark questions (\cref{sec:s-masking}).
We render pages at 144 dpi and cover every line that contains an instance with a gray box. The conditions are no mask
(M0), the gold instances (M\textsubscript{gold}), the copies only, all instances (M\textsubscript{all}), and random
lines of the same area (M\textsubscript{rand}). Each input has six pages (gold, copy and BM25 distractor pages), fixed
across conditions, so the copy is always present. This is common in practice. For half of the
questions with a cross-page copy, the BM25 top-5 contains both the gold page and a copy, and 24\% of \MDR candidate
lists contain a gold and a non-gold quote for the same fact.

Answers are scored with the rule-based scorer of \MLB~\cite{ma2024mmlongbench} and count as correct at a score of at least
0.8. The redundancy effect
\(\dred=P(\text{correct under M}_\text{gold}\wedge\text{wrong under M}_\text{all}\mid\text{correct under M0})\)
is the share of correct answers whose dependence on the fact a single-region test misses. Its denominator includes
correct answers that may not use the gold region. We evaluate Qwen2.5-VL-7B~\cite{bai2025qwen25vl} and
InternVL3-8B~\cite{zhu2025internvl3} on page images with greedy decoding, Qwen3-4B-Instruct-2507~\cite{yang2025qwen3} on
the page text with masked lines replaced by \texttt{[masked]}, and GPT-6.1 Sol on page images. GPT-6.1 ran without
temperature control and serves as a robustness check. The validity and effect criteria, the no-document runs and the
sensitivity to annotation errors are in \cref{sec:s-masking}; all criteria were fixed before running.

\paragraph{Results.}
\Cref{tab:main}(b) gives the main numbers and \cref{tab:s-masking-full} all conditions. With a copy in the input,
76--94\% of the correctly answered copy items survive M\textsubscript{gold}, and 20--53\% survive
M\textsubscript{all}. For 41--56\% of these items, a single-region test declares the gold region unnecessary although
removing every instance of the fact changes the answer. Every lower bound of \dred\ is above 22 points, and every \dred\
exceeds the change caused by random masks of the same area (3.7--10.1\%). Some items survive M\textsubscript{all} because the value is still visible in a
figure or logo, which lowers \dred.

Masking passes the pre-set validity check for Qwen2.5-VL and the text model: 14.0\% and 11.6\% of the controls that
cannot be answered without the document survive M\textsubscript{gold} (criterion: at most 20\%). InternVL3 and GPT-6.1
fail it with 20.4\% and 28.4\%. In 8 of InternVL3's 11 and 14 of GPT-6.1's 19 surviving controls the fact is still in
the input, which our text-based selection of controls missed. Such answers survive both masks and lower \dred. Given
identical inputs, GPT-6.1 changed its correctness on 3\% of controls. On the 81 inputs from our test sample alone,
\dred{} is 40--55\%, with every lower bound above zero (\cref{sec:s-masking}).

\paragraph{Model-cited regions.}
EVICT~\cite{kotte2026evict} masks the region a model cites and abstains when the answer does not change (CED $=0$). For
a correct answer, CED $=0$ is a false abstention. If the cited region is the annotated gold region and an answer counts
as unchanged while it stays correct, the CED $=0$ rate equals M\textsubscript{gold} survival. For Qwen2.5-VL this is
86.2\% on copy items and 15.7\% on controls, so the test would abstain on most
correct answers whose fact has a copy. With the model's own boxes the result is dominated by localisation errors. Only
18\% of Qwen2.5-VL's 99 parsable boxes overlap an instance of the fact (\cref{sec:s-evict}), so we draw no conclusion
about deployed attribution systems.

\paragraph{Ablation-based attribution.}
ContextCite~\cite{cohenwang2024contextcite} fits a sparse linear model to random ablations of the context. We ran it with
the text model on the same inputs (128 ablations over text blocks; \cref{sec:s-baselines}). When a copy is present, the
gold block ranks lower (mean best rank 13.0, against 2.7 on controls) and receives about a third of a control's score.
All instances together receive about as much as the gold block of a control, so the credit is split among the copies.
Block rankings are only moderately stable between 64 and 128 ablations, and we report aggregates only.

\paragraph{Benchmarks built by masking.}
DocAttriBench~\cite{degrandis2026docattribench} builds element-level gold on single gold pages. It masks each layout
element, keeps the one whose removal most raises the perplexity of the answer, and filters ambiguous cases, which its
authors note include information repeated on the page. We matched 273 of the 276 items of its \MLB subset to the source
questions. Of the 145 single-page questions we annotated, it retains 68\% (78/115) of those without a copy and 37\%
(11/30) of those with one. The gap holds within every evidence type (Mantel--Haenszel odds ratio 0.21, $p=0.002$), and
questions with copies fall from 21\% to 12\% of the subset. This is an association. We do not have the filtering logs
and did not check that every question entered its candidate pool. It agrees with the filter, though: for facts with
copies elsewhere, the key string appears in two or more text blocks of the gold page more often (50\% vs.\ 16\%). The
boxes the benchmark keeps are accurate (\cref{sec:s-dab}), but the questions on which single-region tests are least
reliable are under-represented.

\section{Set-level masking with automatic copy detection}
\label{sec:method}

Set-level masking needs $E(u)$, which benchmarks do not provide. We construct it without training.

\paragraph{Pipeline.}
(1) \emph{Fact extraction}: given the region under test (a gold region, or the region a model cited), a VLM reads a crop
of it with the question and the answer under test (here the reference answer) and writes the fact it provides as one
sentence plus the key strings as printed. (2) \emph{Candidates}: text blocks (OCR lines without a text layer) outside
the region under test, on pages that contain a key string or rank in the top 10 by BM25 against the fact. (3)
\emph{Verification}: the same VLM sees each candidate crop and the fact and answers whether the region \emph{by itself}
establishes the same fact with the same value; accepted regions and the region under test form $\hat{E}(u)$. (4)
\emph{Set-level masking}: mask every region in $\hat{E}(u)$ (M\textsubscript{auto}).

\begin{table*}[t]
  \centering\footnotesize
  \setlength{\tabcolsep}{4.2pt}
  \begin{tabular}{@{}llccccccc@{}}
    \toprule
    & & & \multicolumn{3}{c}{Answer survival (\%)} & \multicolumn{3}{c}{Selectivity} \\
    \cmidrule(lr){4-6}\cmidrule(l){7-9}
    Method & Mask & No copy annot. & Qwen2.5-VL & InternVL3 & Qwen3 (text) & Copy prec.\ (\%) $\uparrow$ & Ctrl.\ flips (\%) $\downarrow$ & Area $\downarrow$ \\
    \midrule
    Single-region masking (gold only) & M\textsubscript{gold} & \cmark & 86.2 & 76.9 & 76.0 & -- & -- & -- \\
    Key-string masking, unverified & M\textsubscript{str}\textsuperscript{blk} & \cmark & 20.7 & 19.2 & 8.0 & 27.5 & 9.5 & 3.1$\times$ \\
    \textbf{Ours: verified set masking} & M\textsubscript{auto} & \cmark & 37.9 & 34.6 & 32.0 & \textbf{60.6} & \textbf{3.3} & \textbf{1.0$\times$} \\
    \midrule
    \textit{Reference: annotated sets} & M\textsubscript{all}\textsuperscript{blk} & \xmark & 24.1 & 23.1 & 20.0 & -- & -- & -- \\
    \bottomrule
  \end{tabular}
  \caption{\textbf{Set-level masking: baselines vs.\ ours} (\MLB copy items answered correctly without a mask, six-page
  input, $n=29$, 26 and 25; all masks at block granularity). Single-region masking is the test used by
  DocAttriBench~\cite{degrandis2026docattribench} and EVICT~\cite{kotte2026evict}. Survival should approach the
  annotated reference. Copy prec.: share
  of the blocks masked beyond the gold region that overlap an annotated or adjudicated copy. Ctrl.\ flips: correctly
  answered controls whose correctness differs from single-region masking, mean of the three models. Area: masked area
  relative to ours. Sets are detected with Qwen2.5-VL for every model.}
  \label{tab:method}
\end{table*}

\paragraph{Detection.}
We score detection at page level against our annotation (138 test units with locatable key strings, 57 copy pages);
the pre-set criterion was recall $\ge0.70$ at precision $\ge0.50$. Accepting every candidate that contains a key string
gives recall 0.88 but precision 0.25, since numbers such as ``41\%'' recur with unrelated meanings. The verified detector,
whose candidates cover 77\% of the copy pages, reaches recall 0.47 at precision 0.50; a reviewer adjudicated all 28 accepted pages outside our annotation and found 10
strict copies that our annotation missed, giving precision 0.68 and recall 0.55, still short of the criterion.

\paragraph{Comparison with baselines.}
\Cref{tab:method} compares M\textsubscript{auto} with single-region masking (M\textsubscript{gold}) and with a simpler
baseline that masks every block containing an extracted key string, without verification
(M\textsubscript{str}\textsuperscript{blk}). Masking every annotated instance at block level
(M\textsubscript{all}\textsuperscript{blk}) gives the reference. Masking needs only the copies in the input, and the
detector found one for 21 of the 29 copy items that Qwen2.5-VL answered correctly.

M\textsubscript{auto} lowers survival from 76--86\% to 32--38\% and closes 78\% of the gap to the reference for each of
the three models, with criteria fixed before running (\cref{sec:s-matched}). GPT-6.1, run with M\textsubscript{auto}
only, drops from 93.8\% to 68.8\%. Removing the same amount of unrelated context does not reproduce the effect: adding
random non-evidence blocks of M\textsubscript{auto}'s extra area to M\textsubscript{gold} changes survival by $-4.3$ to
$+4.8$ points. On controls, M\textsubscript{auto} lowers accuracy by at most 2.3 points net relative to
M\textsubscript{gold}.

The string baseline removes the answer as well as the annotated sets do, but it is much less selective. It masks three
times the area of M\textsubscript{auto}, only 27\% of the blocks it adds are annotated copies (61\% for
M\textsubscript{auto}), and it flips the correctness of 9.5\% of correctly answered controls on average (3.3\% for
M\textsubscript{auto}). Among the same key-string candidates, verification raises block
precision from 39\% to 76\%, which matters when the masked set must leave the rest of the input intact.

\section{Discussion and limitations}
\label{sec:discussion}

\paragraph{Size of the effect.}
Copies are common but not universal. They affect about a fifth of long-document questions by our original annotation
and about a third after the review's correction, with more in financial reports and among retrieved candidates. Their
effect on aggregate rankings is moderate (11\% pairwise flips among competitive systems, same top system), but single
systems move by ten or more places, and the original metric favours systems that cite many quotes. For masking-based
attribution the effect on fact dependence is large in all four models.

\paragraph{Recommendations.}
Benchmarks that score evidence should (i) annotate evidence as sets per fact unit, (ii) separate contextual gold from
fact-bearing gold, (iii) score retrieval and citations against the set rather than against answer strings or unchecked LLM judgments, and (iv) not build gold or abstention decisions
on single-region masking without first collecting the region's copies, and report which cases a masking filter
removes.

\paragraph{Limitations.}
One annotator labelled all questions; the independent review finds errors both ways, mostly missed copies. Copy
search relies on the text layer and OCR, so pixel-only copies are under-counted, and \MDR image candidates were judged
partly from their descriptions. The \MLB development and test samples disagree (43\% vs.\ 20\%). Masking uses 25--32
correct copy items per model from a selected set (8 copy items lack item-level verification records), in a setting that contains a copy by construction, and for two models the masking validity
check failed its pre-set criterion. Copy detection misses its pre-set recall
criterion at the document level. We tested two open 7--8B judges; stronger judges may come closer. The DocAttriBench analysis is correlational; CiteVQA is not yet
audited.

\section{Conclusion}
The same fact often appears in several places in a long visual document, and single-gold evaluation treats all but one
of them as wrong. We measured how often this happens in two benchmarks and showed that it distorts retrieval and citation
scores and hides fact-level dependence from single-region masking. Set-level masking with automatically detected
copies recovers most of this dependence. Annotations, metrics, code and
model outputs are provided as ancillary files.

{\small
\bibliographystyle{ieeenat_fullname}
\bibliography{main}
}
\maketitlesupplementary
\label{supp:start}
\setcounter{section}{0}\renewcommand{\thesection}{\Alph{section}}\renewcommand{\theHsection}{supp.\Alph{section}}
\setcounter{table}{0}\renewcommand{\thetable}{S\arabic{table}}\renewcommand{\theHtable}{supp.S\arabic{table}}
\setcounter{figure}{0}\renewcommand{\thefigure}{S\arabic{figure}}\renewcommand{\theHfigure}{supp.S\arabic{figure}}

\noindent Sections A--I, Tables S1--S20 and Figures S1--S2 are in this supplement; all other section, table and figure numbers refer
to the main paper.

\section{Annotation and its independent review}
\label{sec:s-annot}

\paragraph{Samples.}
The \MLB development sample (60 questions) contains answerable questions with one or two gold pages, 10 per stratum
(plain text, table, chart, figure, layout, two-page), one question per document (seed 20261015). It was annotated first
and used to fix the protocol; two of its positives were later downgraded to loose under the final strict rule. The test
sample (160) uses the same strata, excludes the development questions, allows at most two questions per document, and
has quotas of 27 per single-source stratum and 44 two-page questions (the layout stratum yielded 8 under the
per-document cap; seed 20261015). The two samples therefore differ in per-document cap and stratum quotas, but neither
difference explains the higher development rate (two-page questions, which have the highest rate, are a smaller share of
the development sample). \MDR questions are two uniform random samples of the 2{,}000-question evaluation set (seeds
20261015 and 20261016). Unit-level instance annotation exists for the \MLB test sample and all \MDR questions; the \MLB
development sample has question-level labels and page-level units.

\paragraph{Annotator.}
All annotations were produced by a single annotator, an LLM (Claude), following a written protocol: identify the fact
units of the reference answer; verify the gold pages; search the whole document for instances by string and number
search over the text layer (OCR for 13 scanned slide decks), read candidate contexts, and render pages when needed;
label each instance strict, loose, combination or contextual. For \MDR, each of the 15 candidates was read; image
candidates were judged from the source page's text layer where the PDF is available and otherwise from the dataset's
image description. Boxes are located automatically from key strings and are absent where a fact exists only in pixels.
Because the annotator is a model, the human review below is the main check on the labels.

\paragraph{Second search pass.}
For 30 random test questions labelled copy-free, we searched again with BM25 over the fact context and number-format
variants of the key strings; no missed copy was found. This pass shares the text-layer limitation of the annotation and
does not replace the human review.

\paragraph{Independent human review.}
For each benchmark we drew 15 questions labelled redundant and 15 labelled copy-free (seed 20261017). Two human
reviewers, without access to our labels or to each other's, re-annotated them: for \MLB they searched the whole PDF for
regions that by themselves establish each required fact; for \MDR they labelled every candidate quote as a strict
copy, a loose copy, or neither. The acceptance criteria were fixed before any review was returned: Cohen's
$\kappa\ge0.6$ between reviewers on each benchmark (question level, strict copies), and a pooled confirmation of our
positives of at least 11/15 per reviewer. \Cref{tab:s-review} gives the results. Both criteria are met.

\begin{table}[t]
  \centering\footnotesize\setlength{\tabcolsep}{2pt}
  \begin{tabular}{@{}lcc@{}}
    \toprule
    & \MLB & \MDR \\
    \midrule
    $\kappa$ between reviewers & 0.93 & 0.93 \\
    $\kappa$ A / B vs.\ our labels & 0.87 / 0.80 & 0.73 / 0.67 \\
    Positives confirmed (strict) & 30/30 & 26/30 \\
    \quad strict or loose & 30/30 & 30/30 \\
    Copies in negatives: A / B & 2/15 / 3/15 & 2/15 / 3/15 \\
    \quad found by both reviewers & 2 & 2 \\
    Copy pages match ours & 15/15 & -- \\
    \bottomrule
  \end{tabular}
  \caption{\textbf{Independent human review} of 30 questions per benchmark (15 positives, 15 negatives by our
  labels). Not confirmed as strict on \MDR: questions 141 and 1750 (both reviewers: loose). Copies found by both
  reviewers in our negatives: \MLB 404 and 454 (the latter a bar label in a figure, missed by text search); \MDR 1267
  and 1582.}
  \label{tab:s-review}
\end{table}

\paragraph{Corrected rates.}
Writing $r$ for our rate, $c$ for the confirmation rate of our positives and $m$ for the rate at which reviewers found
copies in our negatives, the corrected rate is $r c + (1-r)\,m$ (\cref{tab:s-correct}). The miss rate $m$ rests on 15
questions per benchmark (2/15 has a 95\% Wilson interval of 3.7--37.9\%), so the corrections are imprecise; the main
paper reports the original annotation, frozen before the review, in every table.

\begin{table}[t]
  \centering\footnotesize\setlength{\tabcolsep}{2pt}
  \begin{tabular}{@{}lcccc@{}}
    \toprule
    & $r$ & $c$ & $m$ & corrected \\
    \midrule
    \MLB test, found by both & 19.5 & 1.00 & 0.133 & 30.2 \\
    \quad mean over reviewers & 19.5 & 1.00 & 0.167 & 32.9 \\
    \MDR, found by both & 28.1 & 0.867 & 0.133 & 33.9 \\
    \quad mean over reviewers & 28.1 & 0.867 & 0.167 & 36.3 \\
    \bottomrule
  \end{tabular}
  \caption{\textbf{Review-corrected redundancy rates} (\%).}
  \label{tab:s-correct}
\end{table}

\paragraph{Relation-level agreement.}
We also evaluated the review at the relation level. Inter-reviewer agreement is $\kappa=0.96$ for \MDR (915 candidate
quote $\times$ fact unit relations; 0.97 on non-gold quotes) and $0.91$ for \MLB (fact unit $\times$ page). Taking the
relations judged strict by both reviewers (by either reviewer) as reference, our strict relations have a precision of
0.74 (0.79) and a recall of 0.87 (0.86) on \MDR, and a precision of 0.97 (1.00) and a recall of 0.58 on \MLB. Agreement
and precision both fall in the highest tier (\emph{good}) of the thresholds fixed before the review; the lower \MLB
recall reflects copies missed by text-layer search. For \MLB the relations are the pages nominated by us or by either
reviewer, so $\kappa$ there measures agreement on nominated pages rather than specificity over every page of the PDF.

\section{Test-sample-only analyses}
\label{sec:s-test}
\Cref{fig:prevalence} and the retrieval analysis (\cref{sec:retrieval}) pool the development and test samples of \MLB. \Cref{tab:s-doctype,tab:s-retrieval}
repeat them on the test sample alone. The test sample confirms the pooled picture: financial reports remain the most
redundant document type (9/17, 53\%), and evidence-aware scoring gives higher recall for both retrievers at both cutoffs
(all four gaps positive, three of four intervals excluding zero); as in the pooled sample, the gain concentrates on
questions with copies.

\begin{table}[t]
  \centering\footnotesize\setlength{\tabcolsep}{2pt}
  \begin{tabular}{@{}lcc@{}}
    \toprule
    Document type & Pooled & Test only \\
    \midrule
    Financial report & 14/26 (54\%) & 9/17 (53\%) \\
    Research report & 15/54 (28\%) & 7/38 (18\%) \\
    Academic paper & 10/38 (26\%) & 7/31 (23\%) \\
    Administration/Industry file & 4/13 (31\%) & 3/11 (27\%) \\
    Brochure & 4/22 (18\%) & 3/18 (17\%) \\
    Guidebook & 5/31 (16\%) & 1/22 (5\%) \\
    Tutorial/Workshop & 1/24 (4\%) & 0/17 (0\%) \\
    \midrule
    All & 53/208 (25\%) & 30/154 (19\%) \\
    \bottomrule
  \end{tabular}
  \caption{\textbf{Questions with a strict copy, by document type} (\MLB, applicable questions).}
  \label{tab:s-doctype}
\end{table}

\begin{table}[t]
  \centering\footnotesize\setlength{\tabcolsep}{2pt}
  \begin{tabular}{@{}llcccc@{}}
    \toprule
    & Hit counted on & R@1 & R@5 \\
    \midrule
    \multirow{4}{*}{BM25} & Dataset pages & 36.4 & 66.2 \\
    & Verified gold & 41.6 & 73.4 \\
    & Evidence-aware & 44.8 & 76.6 \\
    & Gap {\scriptsize[95\% CI]} & +3.2 {\scriptsize[0.6, 6.5]} & +3.2 {\scriptsize[0.6, 6.5]} \\
    \midrule
    \multirow{4}{*}{E5} & Dataset pages & 37.7 & 65.6 \\
    & Verified gold & 42.9 & 72.7 \\
    & Evidence-aware & 44.2 & 75.3 \\
    & Gap {\scriptsize[95\% CI]} & +1.3 {\scriptsize[0.0, 3.2]} & +2.6 {\scriptsize[0.6, 5.2]} \\
    \bottomrule
  \end{tabular}
  \caption{\textbf{Page retrieval on the \MLB test sample only} (154 applicable questions; recall, \%). Gap: evidence-aware
  minus verified gold.}
  \label{tab:s-retrieval}
\end{table}

\section{Quote re-scoring}
\label{sec:s-quotes}

\paragraph{Metric variants.}
Let $C$ be the cited quotes, $G$ the gold quotes, and $S(q)$ the set of units quote $q$ supports by itself (strict
copies; loose copies only when stated). A unit is \emph{covered} if some cited quote supports it, or all quotes of one
of its listed combinations are cited. \emph{Original}: $P=|C\cap G|/|C|$, $R=|C\cap G|/|G|$. \emph{A}: precision counts
every quote in $G\cup\{q:S(q)\neq\emptyset\}$; recall as original. \emph{C}: precision as A; a gold quote counts as
recalled if it is cited, or if every unit it supports is covered (or every unit for whose combination it is needed);
contextual gold, which supports no unit, must still be cited. \emph{R}: precision as A; recall is the share of
supported units that are covered, so contextual gold is no longer required. Questions marked not applicable use the
original definition. F1 is averaged over questions, then over systems.

\paragraph{Sensitivity to the review's corrections.}
The re-scoring is robust to the human review. We applied every relation-level judgement of the review to the 30
reviewed \MDR questions: a (quote, unit) relation is strict if both reviewers (or either) marked it strict; all other
questions, the benchmark's gold lists and the combinations are unchanged. This includes the four question-level
corrections of \cref{tab:s-review}. \Cref{tab:s-t7} shows that Quote F1 under C changes by at most 0.003, the flip
rate stays within the bootstrap interval of the original, and the top system is unchanged. On \MLB, the reviewed copy
pages change a single retrieval hit (E5, R@5). The reviewed questions were drawn half from our positives, so this is a
sensitivity analysis rather than a corrected estimate. Both corrected files are released alongside the original.

\begin{table}[t]
  \centering\footnotesize\setlength{\tabcolsep}{2pt}
  \begin{tabular}{@{}lccccc@{}}
    \toprule
    Annotation & C F1 & R F1 & $\tau_{\text{comp}}$ & C flips & Top-1 \\
    \midrule
    Original & .601 & .627 & .771 & 11.4\% {\scriptsize[8.5, 14.8]} & GPT-4.1 \\
    Review, both strict & .598 & .616 & .796 & 10.2\% {\scriptsize[7.5, 13.7]} & GPT-4.1 \\
    Review, either strict & .599 & .622 & .797 & 10.1\% {\scriptsize[7.6, 13.8]} & GPT-4.1 \\
    \bottomrule
  \end{tabular}
  \caption{\textbf{Re-scoring with the human review's relation-level judgements} (91 systems; flips among the 65
  competitive systems, 2{,}080 pairs).}
  \label{tab:s-t7}
\end{table}

\paragraph{Modality of the copies.}
Of the 71 non-gold quotes that strictly support a fact, 45 (63\%) have the other modality than the gold quotes for the
same fact (a text passage restating an image or the reverse); among the 61 whose fact also has a strict gold quote, the
share is 74\%. Weighted by how often systems cite them, at least 63\% [49, 76] of the valid citations counted as wrong
are cross-modal, and in 85\% [80, 90] of the coverage credits of C at least one supporting cited quote has the other
modality than the credited gold quote. On \MLB, 16 of the 30 redundant test questions have a cross-modal copy, most
often a sentence restating a chart (12 gold--copy pairs) or a table (4).

\begin{figure}[t]
  \centering
  \includegraphics[width=\linewidth]{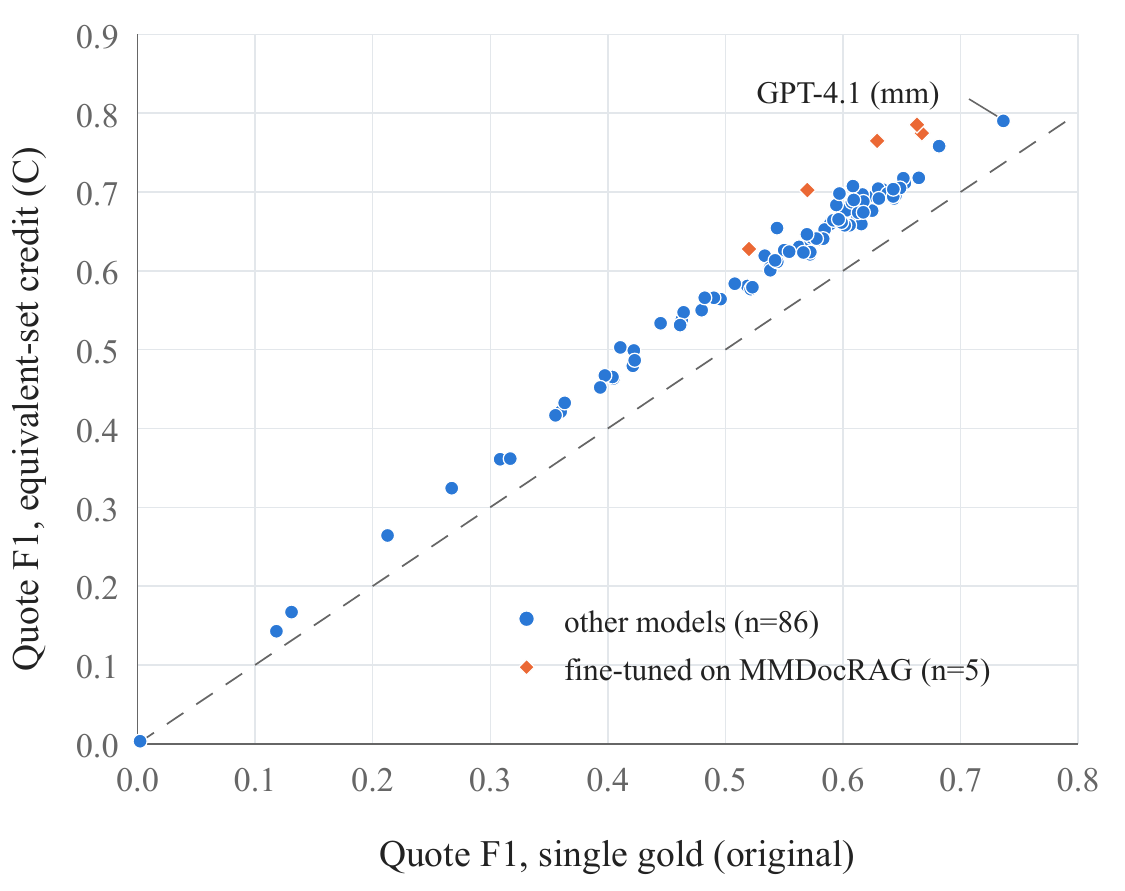}
  \caption{\textbf{Original vs.\ equivalent-set Quote F1} for 91 systems. Systems fine-tuned on \MDR, which cite about
  two quotes per question, gain most.}
  \label{fig:scatter}
\end{figure}

\begin{figure}[t]
  \centering
  \includegraphics[width=\linewidth]{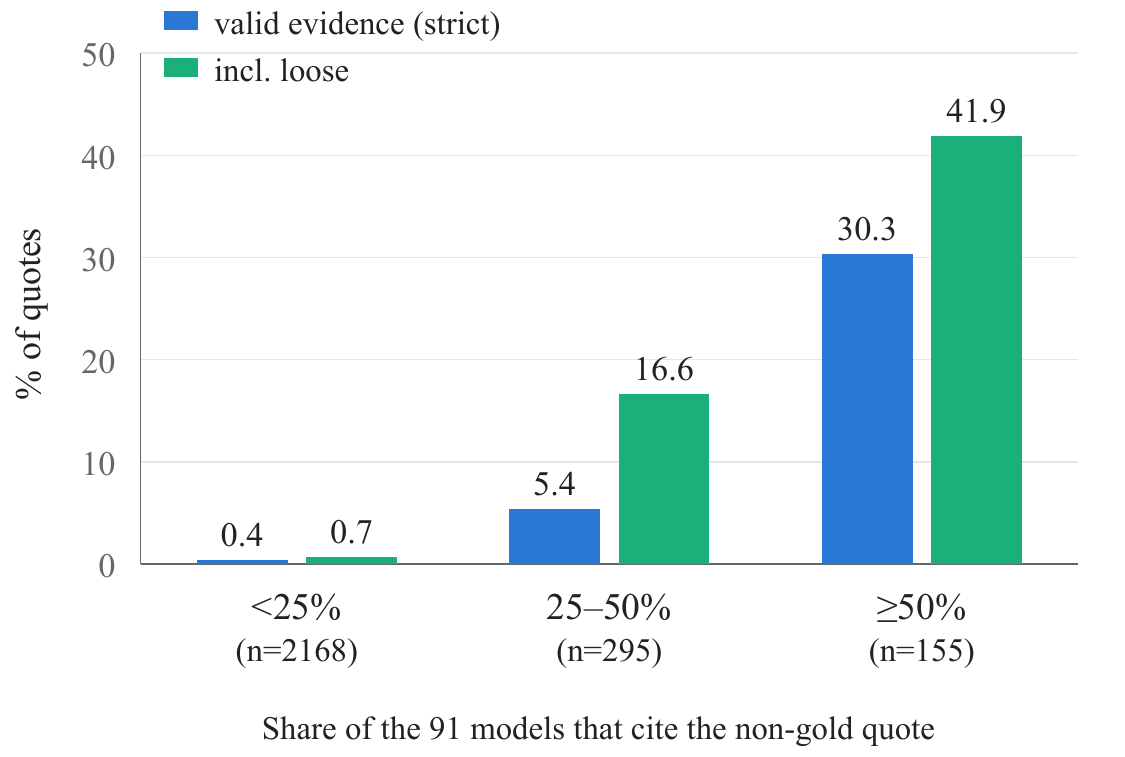}
  \caption{\textbf{Consensus marks unlabelled copies.} Share of non-gold \MDR quotes that are valid evidence, by how many
  of the 91 systems cite them.}
  \label{fig:consensus}
\end{figure}

\paragraph{All systems.}
\Cref{tab:s-systems} lists Quote F1 and rank for all 91 published system outputs under the original metric and under C,
on our 200 questions; \cref{tab:main}(c) in the main paper shows the 12 that rank in the top ten under either metric.

\begin{table*}[t]
  \centering\scriptsize\setlength{\tabcolsep}{2.5pt}
  \begin{tabular}{@{}llcc@{\hspace{10pt}}llcc@{\hspace{10pt}}llcc@{}}
    \toprule
    System & Input & Original & C & System & Input & Original & C & System & Input & Original & C \\
    \midrule
    GPT-4.1 & MM & .736 (1) & .790 (1) & Gemini-2.5-Flash & Text & .606 (32) & .658 (40) & Qwen2.5-3B-FT & Text & .520 (63) & .628 (51) \\
    GPT-4.1 & Text & .682 (2) & .758 (5) & Gemini-2.0-Flash & MM & .603 (33) & .677 (30) & InternVL3-14B & MM & .519 (64) & .581 (63) \\
    Qwen2.5-32B-FT & Text & .667 (3) & .775 (3) & Grok-3 & Text & .601 (34) & .658 (41) & InternVL3-9B & Text & .508 (65) & .584 (62) \\
    Gemini-2.5-Pro & MM & .665 (4) & .718 (6) & Llama-4-Maverick & Text & .599 (35) & .661 (37) & InternVL3-14B & Text & .496 (66) & .564 (68) \\
    Qwen2.5-72B-FT & Text & .663 (5) & .785 (2) & Llama-4-Maverick & MM & .597 (36) & .698 (18) & Qwen-VL-Max & MM & .490 (67) & .566 (67) \\
    Gemini-2.0-Flash-Think & MM & .653 (6) & .711 (8) & Qwen3-14B & Text & .596 (37) & .665 (35) & Qwen2.5-7B & Text & .482 (68) & .566 (66) \\
    GPT-4o & MM & .651 (7) & .718 (7) & GPT-4-Turbo & Text & .594 (38) & .684 (29) & Llama-4-Scout & MM & .480 (69) & .550 (69) \\
    Grok-3-mini & Text & .649 (8) & .705 (10) & GPT-4o-mini & Text & .592 (39) & .664 (36) & GPT-4.1-nano & Text & .464 (70) & .547 (70) \\
    Claude-3.5-Sonnet & MM & .646 (9) & .697 (19) & Gemini-2.0-Flash-Think & Text & .589 (40) & .659 (39) & R1-Distill-Qwen-32B & Text & .463 (71) & .538 (71) \\
    GPT-4.1-mini & MM & .645 (10) & .695 (21) & Gemini-2.0-Pro & OCR & .584 (41) & .652 (43) & Llama-3.1-8B & Text & .461 (72) & .531 (73) \\
    GPT-4o & Text & .644 (11) & .704 (12) & Qwen-Plus & Text & .583 (42) & .641 (47) & InternVL2.5-38B & MM & .445 (73) & .533 (72) \\
    Gemini-2.5-Pro & Text & .643 (12) & .692 (24) & Gemini-1.5-Pro & Text & .580 (43) & .640 (48) & Mistral-Small-24B & Text & .423 (74) & .487 (76) \\
    Qwen3-8B & Text & .643 (13) & .704 (13) & Qwen2.5-VL-72B & MM & .577 (44) & .641 (46) & InternVL2.5-8B & MM & .422 (75) & .499 (75) \\
    Qwen3-30B-A3B & Text & .643 (14) & .694 (23) & Qwen2.5-14B & Text & .573 (45) & .642 (45) & Mistral-7B & Text & .421 (76) & .479 (77) \\
    DeepSeek-V3 & Text & .642 (15) & .703 (15) & InternVL3-38B & Text & .573 (46) & .637 (49) & InternVL2.5-78B & MM & .411 (77) & .503 (74) \\
    Gemini-2.5-Flash & MM & .638 (16) & .698 (17) & Llama-3.3-70B & Text & .572 (47) & .624 (54) & InternVL3-8B & MM & .405 (78) & .463 (80) \\
    Gemini-2.0-Pro & MM & .634 (17) & .703 (14) & InternVL3-78B & Text & .572 (48) & .621 (56) & Llama-3.1-8B & OCR & .404 (79) & .465 (79) \\
    GPT-4.1-mini & Text & .631 (18) & .692 (25) & Qwen2.5-7B-FT & Text & .570 (49) & .702 (16) & Qwen2.5-7B & OCR & .398 (80) & .467 (78) \\
    Gemini-2.0-Pro & Text & .630 (19) & .704 (11) & Qwen3-4B & Text & .569 (50) & .646 (44) & Qwen2.5-VL-32B & MM & .394 (81) & .452 (81) \\
    Qwen2.5-14B-FT & Text & .629 (20) & .765 (4) & Qwen3-32B & Text & .566 (51) & .623 (55) & Mixtral-8x7B & Text & .363 (82) & .432 (82) \\
    Gemini-1.5-Pro & MM & .627 (21) & .695 (22) & GPT-4o & OCR & .563 (52) & .630 (50) & GPT-4.1-nano & MM & .360 (83) & .421 (83) \\
    DeepSeek-R1 & Text & .625 (22) & .676 (31) & InternVL3-38B & MM & .554 (53) & .624 (53) & InternVL2.5-26B & MM & .355 (84) & .417 (84) \\
    Qwen-Max & Text & .617 (23) & .688 (27) & Gemini-2.0-Flash-Think & OCR & .550 (54) & .626 (52) & Qwen2.5-3B & Text & .317 (85) & .362 (85) \\
    Qwen3-235B-A22B & Text & .617 (24) & .674 (32) & InternVL3-8B & Text & .544 (55) & .611 (59) & Llama-3.2-3B & Text & .308 (86) & .361 (86) \\
    o3-mini & Text & .617 (25) & .697 (20) & InternVL3-9B & MM & .544 (56) & .654 (42) & Qwen-VL-Plus & MM & .267 (87) & .324 (87) \\
    Claude-3.5-Sonnet & Text & .616 (26) & .659 (38) & Qwen2.5-72B & OCR & .542 (57) & .613 (58) & Qwen2.5-VL-7B & MM & .213 (88) & .264 (88) \\
    QwQ-Plus & Text & .613 (27) & .674 (33) & Gemini-2.0-Flash & Text & .538 (58) & .601 (61) & MiniCPM-o-2.6 & MM & .131 (89) & .167 (89) \\
    InternVL3-78B & MM & .612 (28) & .670 (34) & DeepSeek-V3 & OCR & .538 (59) & .604 (60) & QVQ-Max & MM & .118 (90) & .143 (90) \\
    Qwen2.5-72B & Text & .609 (29) & .690 (26) & R1-Distill-Llama-70B & Text & .534 (60) & .619 (57) & Janus-Pro-7B & MM & .002 (91) & .003 (91) \\
    Qwen2.5-32B & Text & .608 (30) & .708 (9) & Llama-4-Scout & Text & .523 (61) & .579 (64) & & & & \\
    GPT-4o-mini & MM & .608 (31) & .687 (28) & Llama-3.3-70B & OCR & .522 (62) & .577 (65) & & & & \\
    \bottomrule
  \end{tabular}
  \caption{\textbf{Quote F1 (rank) of all 91 \MDR systems} on our 200 questions, sorted by the original metric. MM:
  images given as images; Text and OCR: the benchmark's two text-only input modes. FT: fine-tuned by the \MDR authors.}
  \label{tab:s-systems}
\end{table*}

\paragraph{Stability of the C flip rate.}
\Cref{tab:s-flips} reports a question-level bootstrap of the flip rate among the 65 systems with original F1 $\ge0.5$
(2{,}000 resamples, seed 20261015; the system set is held fixed), the share of flipped pairs that were nearly tied under
the original metric, and the flip rate under other thresholds for the competitive set. The flip rate is stable
under resampling (95\% CI 8.5--14.8\%) and grows among the strongest systems, where rankings matter most: from 8.0\%
with 79 systems to 18.4\% with the 34 systems above 0.6. Most flipped pairs (59\%) were separated by at least 0.01 F1
under the original metric, so the correction changes orderings the original metric did resolve. A flip is a pair whose F1
differences under the two metrics have strictly opposite signs.

\begin{table}[t]
  \centering\footnotesize\setlength{\tabcolsep}{2pt}
  \begin{tabular}{@{}lcc@{}}
    \toprule
    & Systems & C flip rate \\
    \midrule
    Threshold 0.4 & 79 & 8.0\% (247/3{,}081) \\
    Threshold 0.5 (main paper) & 65 & 11.4\% (238/2{,}080) [8.5, 14.8] \\
    Threshold 0.6 & 34 & 18.4\% (103/561) \\
    \midrule
    \multicolumn{3}{@{}l}{Flipped pairs (threshold 0.5) with original F1 difference} \\
    \quad $\ge0.01$ & \multicolumn{2}{c}{140/238 (58.8\%)} \\
    \quad $\ge0.02$ & \multicolumn{2}{c}{76/238 (31.9\%)} \\
    \bottomrule
  \end{tabular}
  \caption{\textbf{Stability of the C re-ranking.} Interval: 95\% question-level bootstrap.}
  \label{tab:s-flips}
\end{table}

\section{Masking experiments}
\label{sec:s-masking}

\paragraph{Items and inputs.}
The 157 units form a set selected for the masking experiments. It is not a random sample of \MLB and not a subset of
the 160-question test sample. It has 59 units with copies (56 cross-page, 3 same-page) and 98 controls whose
key string occurs only in the gold region. Of the 56 cross-page copy items, 29 come from our test sample, 19 from the
development sample and 8 from \MLB questions outside the 220 we audited; the 3 same-page items come from the
development sample. Of the 98 controls, 52 come from the test sample and 46 from questions outside the audit. The
question text of every item matches the benchmark. Items outside the audit enter neither the prevalence estimates nor
the independent review. The 8 copy items outside the audit come from an earlier answer-string candidate search that
predates the VLM detector of \cref{sec:method}, so \Mx{auto} did not produce them; the detector accepts the selected
copy page for 4 of them. Item-level records of how their copies were checked were not kept. The six-page input (S\_page) contains the gold page(s), the copy page(s) and
distractor pages ranked by BM25 against the question, fixed per item across conditions; it is used for the 56
cross-page copy items and all controls (154 items). The gold-page-only input (S\_single) is built for the 3 same-page
copy items and the controls. Every item is also asked without the document (NoDoc).

\paragraph{Masks.}
Pages are rendered at 144 dpi. Instances are located in the text layer (OCR lines for pages without one; a drawn box
for charts without text) and every line that contains an instance is covered by a gray box padded by 2\,pt.
\Mx{gold} covers gold instances; \Mx{copy} covers copies; \Mx{all} covers gold instances, copies, and any line on a
distractor page that contains a key string but was not verified as a copy, so that no textual occurrence of the fact
remains (such lines occur in 8 of the 56 copy items and in no control; excluding these 8 items changes \dred{} by
$-0.6$ to $+3.8$ points). \Mx{rand} covers random text lines that do not contain
a key string, on the same pages and with the same total area per page as \Mx{all}. \Mx{auto} covers whole text blocks
(OCR lines for pages without a text layer) accepted by the detector (\cref{sec:s-detect}). For the text model, every
line whose box overlaps a mask by at least 30\% of its area is replaced by the token \texttt{[masked]}, and runs of
masked lines are collapsed.

\paragraph{Prompts.}
Image input, system prompt: ``You are given images of pages from one document, in page order. Some regions may be
covered by gray boxes. Answer the question using only the document. If the document does not contain the answer, reply
exactly: Not answerable.'' Text input: the same, with ``the text of pages'' and ``Some lines may be replaced by
[masked]'', and pages separated by ``--- Page $k$ ---''. NoDoc: ``Answer the question. If you cannot answer it, reply
exactly: Not answerable.'' The user turn is the question followed by a format instruction chosen by the dataset's
answer type: ``Answer with a single integer only.'' (Int); ``Answer with a single number only (keep the \% sign if
the answer is a percentage).'' (Float); ``Answer with a short phrase only.'' (Str, None); ``Answer with a Python list
of short strings only, e.g.\ ["a", "b"].'' (List). Output is capped at 96 tokens.

\paragraph{Scoring.}
The answer is the \texttt{answer} field if the output is JSON, otherwise the first line, after removing reasoning tags,
an ``Answer:'' prefix and a final period. It is scored with \MLB's official rule-based scorer: Int by exact match;
Float with percentage rescaling and 1\% relative tolerance; Str by exact match for short or numeric strings and ANLS
(threshold 0.5) otherwise; List element-wise after sorting. An answer is correct if its score is at least 0.8. No model
judge is used.

\paragraph{Models.}
\emph{Qwen2.5-VL-7B-Instruct}: vLLM 0.8.5.post1, transformers 4.51.3, greedy decoding, at most $1280\cdot28\cdot28$
pixels per page. \emph{Qwen3-4B-Instruct-2507}: same stack, greedy; all weight shards and inference files were verified
by SHA-256 against the official release (revision cdbee75f). \emph{InternVL3-8B}: revision 0908e0d6, same vLLM, temperature
0, up to 6 tiles of 448\,px plus a thumbnail per page (comparable to the Qwen2.5-VL resolution), maximum length 16{,}384
tokens; it reused the rendered images and prompts of the first run unchanged (SHA-256 verified). \emph{GPT-6.1 Sol}
(model string \texttt{gpt-6.1-sol}): queried on 2026-10-17 through the Codex app-server protocol at low reasoning effort
and the default service tier, eight jobs in parallel. Each job ran in a fresh process and an ephemeral thread with no
history; images were attached unchanged at original detail; no tools, temperature or output schema were supplied, and
configured external tool servers were disabled. An audit of all raw events found four built-in resource listings that
returned empty results, in two no-document jobs, and no other tool events. All 983 jobs were answered; one job answered
twice after an overlapping retry, and the scorer used the later answer without regard to correctness. An earlier
configuration with tools enabled attempted to fetch a source PDF; all of its outputs were excluded. The interface
exposes no temperature and no immutable model checkpoint.

\paragraph{Pre-set criteria.}
The criteria were fixed on 2026-10-16, before any model output was produced, and revised once on 2026-10-17 (detection
criterion only) before any result was seen; the revision did not change any verdict. \emph{V1} (masking removes the
evidence): among controls not answerable without the document, at most 20\% survive \Mx{gold}. \emph{V2} (no generic
masking artefact): at most 15\% of answers change under \Mx{rand}, and \dred{} minus that change rate has a positive
lower bound. \emph{E1} (effect): survival under \Mx{gold} minus survival under \Mx{all} on copy items, and \dred, have
positive lower bounds. If V1 or V2 fails, the masking instrument is ``not validated''. \emph{B1}--\emph{B3}
(automatic sets): \Mx{auto} lowers survival relative to \Mx{gold} with a positive lower bound; it closes at least half of
the \Mx{gold}--\Mx{all} gap; its answer-change rate on controls, relative to M0, exceeds that of \Mx{gold} by at most 5 points, \ie it lowers
control accuracy by at most 5 points net. Intervals are 95\% cluster
bootstraps over questions (2{,}000 resamples, seed 20261015). \Cref{tab:s-criteria} gives every value.

\paragraph{Full results.}
\Cref{tab:s-masking-full} gives every masking condition for copy items and controls; \cref{tab:main}(b) in the main
paper shows a subset of its columns.

\begin{table}[t]
  \centering
  \footnotesize
  \setlength{\tabcolsep}{2pt}
  \begin{tabular}{@{}llccccc@{}}
    \toprule
    Model & Items ($n$) & M\textsubscript{gold} & M\textsubscript{copy} & M\textsubscript{all} & M\textsubscript{rand} & \dred\ {\scriptsize[95\% CI]} \\
    \midrule
    \multirow{2}{*}{Qwen2.5-VL-7B} & copy (29) & 86.2 & 93.1 & 31.0 & 96.6 & 55.2 {\scriptsize[36.7, 75.0]} \\
     & control (51) & 15.7 & n/a & 15.7 & 96.1 & -- \\
    \midrule
    \multirow{2}{*}{InternVL3-8B} & copy (26) & 76.9 & 92.3 & 30.8 & 96.2 & 46.2 {\scriptsize[26.9, 66.7]} \\
     & control (54) & 20.4 & n/a & 20.4 & 90.7 & -- \\
    \midrule
    \multirow{2}{*}{GPT-6.1 Sol} & copy (32) & 93.8 & 96.9 & 53.1 & 96.9 & 40.6 {\scriptsize[22.6, 59.4]} \\
     & control (75) & 33.3 & n/a & 37.3 & 96.0 & -- \\
    \midrule
    \multirow{2}{*}{Qwen3-4B (text)} & copy (25) & 76.0 & 76.0 & 20.0 & 88.0 & 56.0 {\scriptsize[37.5, 76.0]} \\
     & control (44) & 11.4 & n/a & 11.4 & 90.9 & -- \\
    \bottomrule
  \end{tabular}
  \caption{\textbf{Answer survival under masking} (\% of items answered correctly without a mask, six-page input;
  $n$ of 56 copy and 98 control items). The first three models read page images; Qwen3-4B-Instruct-2507 reads the page
  text. Intervals: cluster bootstrap over questions.}
  \label{tab:s-masking-full}
\end{table}

\paragraph{Test-sample subset.}
Because 73 of the 154 masking inputs lie outside our independent test sample, we re-scored the existing outputs on the
81 inputs that come from it (65 questions; 29 copy items and 52 controls), without running any model again
(\cref{tab:s-audit-test}). The effect holds on this subset. \dred{} is 40.0--54.5\%, in the same range as on all
inputs, and its lower bound is above zero for all four models. \Mx{auto} still closes most or all of the gap to the
annotated block mask. The subset is small, so the intervals are wide, and its controls give a weaker validity check:
only the text model meets the pre-set criterion V1. Removing also the 19 inputs in the detector's development split
leaves 62 inputs, on which \dred{} is 33.3--44.4\% and the text model's interval reaches zero. This analysis was not
planned before the experiment.

\begin{table}[t]
  \centering\footnotesize\setlength{\tabcolsep}{1.2pt}
  \begin{tabular}{@{}lcccc@{}}
    \toprule
    Model & Copy ($n$) & \dred{} {\scriptsize[95\% CI]} & Closed {\scriptsize[95\% CI]} & V1 \\
    \midrule
    Qwen2.5-VL-7B & 11 & 54.5 {\scriptsize[25.0, 90.0]} & 100.0 {\scriptsize[50, 150]} & 24.0 \\
    InternVL3-8B & 13 & 53.8 {\scriptsize[26.7, 83.3]} & 87.5 {\scriptsize[50, 129]} & 34.6 \\
    GPT-6.1 Sol & 14 & 42.9 {\scriptsize[14.3, 71.4]} & -- & 30.3 \\
    Qwen3-4B (text) & 10 & 40.0 {\scriptsize[10.0, 77.8]} & 125.0 {\scriptsize[100, 180]} & 19.0 \\
    \midrule
    All 154 inputs & 25--32 & 40.6--56.0 & 77.8--78.6 & 11.6--28.4 \\
    \bottomrule
  \end{tabular}
  \caption{\textbf{Masking on the 81 test-sample inputs} (post hoc, existing outputs, \%). Copy: correctly answered
  copy items. Closed: share of the \Mx{gold}--\Mx{all}\textsuperscript{blk} gap closed by \Mx{auto}; above 100
  means lower survival under \Mx{auto} than under the annotated block mask, not better detection. V1: survival under
  \Mx{gold} of controls not answerable without the document (pre-set criterion $\le20$). Intervals: question-cluster
  bootstrap. --: \Mx{all}\textsuperscript{blk} not run for GPT-6.1 (line-level gap closed 66.7).}
  \label{tab:s-audit-test}
\end{table}

\begin{table*}[t]
  \centering\footnotesize\setlength{\tabcolsep}{3.5pt}
  \begin{tabular}{@{}lcccccccc@{}}
    \toprule
    Model & M0 acc. & NoDoc & V1 & V2: \Mx{rand} chg. & V2: \dred$-$chg. & E1: \dred & B1: \Mx{gold}$-$\Mx{auto} & B2 / B3 \\
    \midrule
    Qwen2.5-VL-7B & 51.9 & 0.6 & 14.0 \checkmark & 3.7 & 51.7 [33.3, 70.0] & 55.2 [36.7, 75.0] & 48.3 (lb 28.6) & 87.5\% / +2.0 \\
    InternVL3-8B & 51.9 & 0.0 & 20.4 $\times$ & 7.5 & 42.3 [24.0, 64.0] & 46.2 [26.9, 66.7] & 42.3 (lb 24.0) & 91.7\% / $-$1.9 \\
    GPT-6.1 Sol & 69.5 & 8.9 & 28.4 $\times$ & 3.7 & 37.5 [16.1, 58.1] & 40.6 [22.6, 59.4] & 25.0 (lb 9.7) & 61.5\% / $-$5.3 \\
    Qwen3-4B (text) & 44.8 & 0.6 & 11.6 \checkmark & 10.1 & 44.0 [22.2, 66.7] & 56.0 [37.5, 76.0] & 44.0 (lb 20.8) & 78.6\% / +2.3 \\
    \bottomrule
  \end{tabular}
  \caption{\textbf{Every pre-set criterion, per model} (six-page input, \%; accuracies over 154 items). V1 fails for
  InternVL3 and GPT-6.1, so by the pre-set rule their masking is not validated; \cref{tab:s-v1} explains the failures.
  B2: share of the \Mx{gold}--\Mx{all} gap closed by \Mx{auto}; B3: net loss of control accuracy under \Mx{auto} relative to \Mx{gold} (points; negative values are gains). Flip rates,
  which count changes in both directions, are in \cref{tab:s-string}.
  Excluding copy items answerable without the document leaves the results unchanged except for GPT-6.1 ($n=28$,
  \dred{} 42.9 [25.0, 63.0]).}
  \label{tab:s-criteria}
\end{table*}

\paragraph{Why V1 fails for two models.}
We read every control that survives \Mx{gold} and is not answerable without the document (\cref{tab:s-v1}). Most are
redundancy that our text-based selection of controls missed: the fact restated on another page, visible in a picture
that line masks leave intact, or recoverable from remaining content. The two GPT-6.1 survivors of question 454 survive
through a bar label in a figure that the human review later identified as a copy. Dropping the controls whose fact is
still in the input (post hoc) gives V1 values of 6.5\% (3/46) for InternVL3 and 9.4\% (5/53) for GPT-6.1. Such controls
survive both masks, so they cannot raise \dred.

\begin{table}[t]
  \centering\footnotesize\setlength{\tabcolsep}{2pt}
  \begin{tabular}{@{}lcc@{}}
    \toprule
    Reason & InternVL3 & GPT-6.1 \\
    \midrule
    Restated elsewhere in the input & 3 & 4 \\
    Readable from a picture & 3 & 5 \\
    Countable or inferable from the rest & 2 & 3 \\
    Copy found later by the review & 0 & 2 \\
    Guess among few options & 3 & 0 \\
    Recall of the public source only & 0 & 5 \\
    \midrule
    Total & 11 & 19 \\
    \bottomrule
  \end{tabular}
  \caption{\textbf{Controls that survive \Mx{gold}}, excluding controls answerable without the document.}
  \label{tab:s-v1}
\end{table}

\paragraph{Manual checks (Qwen2.5-VL).}
We read all 16 items that make up \dred: each answer under \Mx{gold} is correct and each under \Mx{all} is genuinely
wrong (\eg 92 $\to$ 72), not a scoring artefact. Of the 9 copy items that survive \Mx{all}, most do so because the value
still appears in pixels (a logo, chart or picture) or a long title is repeated elsewhere; one is a lenient match by the
official scorer (6.31 vs.\ 6.29). The model rarely answers ``Not answerable'' (5 of 98 masked controls), so answer
change tracks evidence removal.

\paragraph{Copies found after the experiment.}
Copies found later by the human review or by the detector adjudication (\cref{sec:s-detect}) were not added to
\Mx{all}. Mapped by question and fact, they touch 4 of the 157 items (q189\_x0, q454\_ctrl, q454\_x0, q946\_auto; the
H1c copy of question 404 belongs to no masking item). For q189\_x0 the new copy page is not part of the six-page input,
so only three items actually had an unmasked copy in view. The results are robust to these additions: excluding
the four items changes \dred{} by $-0.6$ to $+4.9$ points and every model keeps a large effect (\cref{tab:s-later}).
The missed copies and the unverified lines masked in 8 items (see \emph{Masks}) bias \dred{} in opposite directions;
excluding either set changes it by at most 5 points. V1 is essentially unchanged (InternVL3 21.2\%, GPT-6.1 26.2\%).

\begin{table}[t]
  \centering\footnotesize\setlength{\tabcolsep}{2pt}
  \begin{tabular}{@{}lccc@{}}
    \toprule
    Model & \dred{} (all) & \dred{} (excluding) & Change \\
    \midrule
    Qwen2.5-VL-7B & 55.2 & 55.6 (15/27) & +0.4 \\
    InternVL3-8B & 46.2 & 45.8 (11/24) & $-$0.3 \\
    GPT-6.1 Sol & 40.6 & 40.0 (12/30) & $-$0.6 \\
    Qwen3-4B (text) & 56.0 & 60.9 (14/23) & +4.9 \\
    \bottomrule
  \end{tabular}
  \caption{\textbf{Sensitivity to copies found after the experiment} (\%, six-page input).}
  \label{tab:s-later}
\end{table}

\paragraph{How often a copy accompanies the gold in practice.}
\Cref{tab:s-cooc} uses the BM25 and E5 rankings of \cref{sec:retrieval} on the 48 \MLB questions with a cross-page copy
(two development questions in this set have only loose copies; restricting to the 46 with strict copies changes no
entry by more than 2 points). Already at $k=5$, about half of these questions retrieve both the gold page and a copy of
the same fact, which is the situation of our six-page input. In \MDR's own 15-candidate inputs, 102 of 196 questions
have a fact supported by two or more candidates, and in 47 (24\%) a gold and a non-gold candidate support the same fact.

\begin{table}[t]
  \centering\footnotesize\setlength{\tabcolsep}{3pt}
  \begin{tabular}{@{}lcccc@{}}
    \toprule
    & \multicolumn{2}{c}{Gold and copy both retrieved} & \multicolumn{2}{c}{Copy without gold} \\
    \cmidrule(lr){2-3}\cmidrule(l){4-5}
    Top-$k$ & BM25 & E5 & BM25 & E5 \\
    \midrule
    1 & 0.0 & 0.0 & 22.9 & 20.8 \\
    3 & 33.3 & 31.2 & 12.5 & 8.3 \\
    5 & 50.0 & 45.8 & 10.4 & 10.4 \\
    10 & 56.2 & 56.2 & 10.4 & 10.4 \\
    \bottomrule
  \end{tabular}
  \caption{\textbf{Gold and copy in retrieved pages} (\% of the 48 \MLB questions with a cross-page copy). ``Both'':
  the top-$k$ pages contain a gold page and a copy page of the same fact.}
  \label{tab:s-cooc}
\end{table}

\paragraph{Gold-page-only input.}
S\_single contains only 3 copy items (all same-page), too few to analyse. On controls, survival under \Mx{gold} is 15.8\%
(Qwen2.5-VL, $n=57$), 13.8\% (InternVL3, $n=58$) and 6.5\% (text model, $n=46$); GPT-6.1 was not run on this input.

\section{EVICT replication}
\label{sec:s-evict}
EVICT~\cite{kotte2026evict} computes $\mathrm{CED}=\mathbb{1}[s<0.7]\,(1-s)$, where $s$ is the token Jaccard similarity
between the answers before and after masking the cited region, and abstains when CED $=0$. For model-cited regions,
Qwen2.5-VL-7B received the image prompt of \cref{sec:s-masking} followed by ``Then point to the single region of the
document that your answer relies on. Reply in JSON only: \{"answer": "...", "image": $\langle$1-based index of the page
image$\rangle$, "bbox\_2d": [x1, y1, x2, y2]\} with the box in pixel coordinates of that image.'' (at most 160 tokens).
We gray-fill the box and re-ask, or the box together with every instance of the fact. For the gold-region version we
count an answer as unchanged when it is still correct, so its rates equal the \Mx{gold} survival of \cref{tab:s-masking-full}.
\Cref{tab:s-evict} gives both versions.

\begin{table}[t]
  \centering\footnotesize\setlength{\tabcolsep}{2pt}
  \begin{tabular}{@{}lccc@{}}
    \toprule
    CED $=0$ among correct answers & Copy & Control & Difference \\
    \midrule
    Gold region (still correct) & 86.2 & 15.7 & 70.5 [52.8, 85.3] \\
    Cited box (Jaccard $\ge0.7$) & 89.5 & 69.2 & 20.2 [$-$1.8, 40.9] \\
    Cited box + all instances & 26.3 & 7.7 & -- \\
    \midrule
    $n$ (gold / cited) & 29 / 19 & 51 / 26 & \\
    \bottomrule
  \end{tabular}
  \caption{\textbf{False abstention of the EVICT test} (\%). Cited boxes: 99 of 154 outputs parsable, 18\% of them
  overlap an instance of the fact. Masking every instance with the cited box lowers CED $=0$ on copy items by 63.2 points
  [42.1, 84.2].}
  \label{tab:s-evict}
\end{table}

\section{Automatic copy detection}
\label{sec:s-detect}

\paragraph{Pipeline details.}
\emph{Fact extraction}: the VLM sees a crop of the region under test (padded by 48\,pt) and the prompt ``This image is a
region of a document page. It is cited as evidence for the question: ``$\langle$question$\rangle$'' (reference answer:
$\langle$answer$\rangle$). State the specific fact that this region provides for answering the question, as one short
sentence, and list the key strings (numbers, names, short phrases) exactly as they are written in the region. Reply in
JSON only: \{"fact": "...", "keys": ["...", "..."]\}''. \emph{Candidates}: pages whose text contains a key string (up
to 40) together with the top 10 pages by BM25 against the fact and key strings; on these pages, every text block outside
the region under test, keeping up to 25 blocks that contain a key string and up to 8 others ranked by BM25 against the
fact. \emph{Verification}: for each candidate crop (padded by 36\,pt), ``Claim: $\langle$fact$\rangle$. Does this image
region by itself state this same claim, with the same entity and the same value(s)? Answer yes only if the region alone
is enough to establish the claim. Answer yes or no.'' (at most 4 tokens). The VLM is Qwen2.5-VL-7B throughout.

\paragraph{Evaluation.}
Detection is scored at page level against our annotation of cross-page copies. Over all 138 test units with locatable
key strings, candidate recall is 0.77, recall 0.47 and precision 0.50 (\cref{sec:method}); over the 127 units with at
least one candidate, the same quantities are 0.80, 0.49 and 0.50, and 15 units have at least one false alarm. On the
development split (95 units with candidates) they are 0.66, 0.38 and 0.51. The detector uses one frozen prompt on both
splits, and the experiment package records no prompt tuning on the development split. This split is not the audit
split: 92 of the 154 masking inputs carry its development label, including 19 of the 81 test-sample inputs.

\paragraph{False-alarm adjudication.}
A third reviewer judged all 28 accepted test pages outside our annotation, seeing each accepted region enlarged
together with the full page and the gold page, and labelled it a strict copy, a loose copy or not a copy. Ten are strict
copies that our text-based annotation missed and 18 are not copies; counting the ten as correct gives precision 0.68
and recall 0.55.

\paragraph{Matched-granularity comparison.}
\label{sec:s-matched}
\Mx{auto} masks whole text blocks while \Mx{all} masks lines, so we repeated the comparison at block granularity on
the six-page input with Qwen2.5-VL, InternVL3 and the text model (1{,}260 new answers; same prompts and settings). The
gates were fixed before any output existed (2026-10-18 02:39 UTC; an amendment adding \Mx{ctx} and G4 at 03:03 UTC,
before the run at 03:29--03:43 UTC). \emph{G1}: \Mx{auto} closes $\ge50\%$ of the gap between \Mx{gold} and
\Mx{all}\textsuperscript{blk}, which masks every block containing an annotated instance. \emph{G3}: random non-key
blocks of \Mx{all}\textsuperscript{blk}'s area change $\le15\%$ of answers. \emph{G4}: \Mx{ctx}, which adds to \Mx{gold}
random non-evidence blocks with the area that \Mx{auto} adds (matched within $\pm5\%$), lowers survival by $<10$ points
relative to \Mx{gold}. A further gate (G2) masks the gold region with the detector's false-positive blocks only; just one
correctly answered item per model has such blocks, so it is uninformative (survival change 0, $-$100 and 0 points) and
we draw no conclusion from it. \Cref{tab:s-matched} gives the results: all three models pass G1, G3 and G4.

\begin{table}[t]
  \centering\footnotesize\setlength{\tabcolsep}{1.5pt}
  \begin{tabular}{@{}lccc@{}}
    \toprule
    & Qwen2.5-VL & InternVL3 & Text model \\
    \midrule
    Survival, \Mx{gold} & 86.2 & 76.9 & 76.0 \\
    Survival, \Mx{auto} & 37.9 & 34.6 & 32.0 \\
    Survival, \Mx{all}\textsuperscript{blk} & 24.1 & 23.1 & 20.0 \\
    G1: gap closed & 78 {\scriptsize[50, 100]} & 79 {\scriptsize[50, 108]} & 79 {\scriptsize[46, 107]} \\
    G4: \Mx{gold}$-$\Mx{ctx} & 4.3 {\scriptsize[0.0, 13.6]} & $-$4.8 {\scriptsize[$-$15, 0]} & $-$4.8 {\scriptsize[$-$20, 10]} \\
    G3: random change & 1.3 {\scriptsize[0.0, 4.0]} & 1.3 {\scriptsize[0.0, 4.2]} & 4.4 {\scriptsize[0.0, 9.9]} \\
    $n$ (G1 / G4 / G3) & 29 / 23 / 78 & 26 / 21 / 77 & 25 / 21 / 68 \\
    \bottomrule
  \end{tabular}
  \caption{\textbf{Matched-granularity masking} (survival and differences in \%, six-page input; intervals: cluster
  bootstrap over questions). G4 is computed on items whose automatic set adds area beyond the gold region; G3 on
  correctly answered items with an area-matched random mask.}
  \label{tab:s-matched}
\end{table}

\section{DocAttriBench analysis}
\label{sec:s-dab}
We used the \texttt{mmlongbenchdoc} subset of DocAttriBench~\cite{degrandis2026docattribench} (276 items, HuggingFace
revision 184aecba). 273 items match a unique \MLB question by document and question text; the item's image is the
question's evidence page. Among the 145 single-page applicable questions we annotated, \cref{tab:s-dab} gives retention
by evidence type. The direction holds in every type and in both samples (development 5/12 vs.\ 15/21, test 6/18 vs.\
63/94). Stratified by evidence type, the Mantel--Haenszel odds ratio is 0.21 (CMH $p=0.002$); unstratified, Fisher
$p=0.003$. On test units with a key string ($n=74$), the key string appears in two or more text blocks of the gold page
for 50\% of units with copies and 16\% of units without. Of the 45 retained items whose box we could compare with our
gold region, 40 boxes cover it; four of the other five answers are lists or counts that need several elements, and in
the fifth the box covers an adjacent line that states the same answer. We do not have the benchmark's filtering logs, so
these are associations.

\begin{table}[t]
  \centering\footnotesize\setlength{\tabcolsep}{2pt}
  \begin{tabular}{@{}lcc@{}}
    \toprule
    Evidence type & With copy & Without copy \\
    \midrule
    Chart & 3/9 & 17/24 \\
    Figure & 0/2 & 20/32 \\
    Layout text & 3/5 & 10/11 \\
    Plain text & 4/6 & 18/25 \\
    Table & 1/8 & 13/23 \\
    \midrule
    All & 11/30 (37\%) & 78/115 (68\%) \\
    \bottomrule
  \end{tabular}
  \caption{\textbf{Questions retained by DocAttriBench}, among the single-page \MLB questions we annotated.}
  \label{tab:s-dab}
\end{table}

\section{Baselines}
\label{sec:s-baselines}
These comparisons were added after the main study. The decision rules for the visual retrievers, the judges and string
masking were written down before any model was called and revised once, also before any model call (the sample used for
the retriever rule; an area-matched control for string masking); both versions are released with their hashes.
Visual retrievers: ColPali v1.3 and ColQwen2 v1.0 with colpali-engine 0.3.10 and its default preprocessing, pages
rendered at 144 dpi, original questions. Judges: Qwen2.5-VL-7B-Instruct and InternVL3-8B, greedy decoding, at most 4
tokens. String masking reuses the pages, prompts, blocks and settings of the matched-granularity experiment
(\cref{sec:s-matched}). Intervals: question-level bootstrap as elsewhere.

\paragraph{Retrieval.}
\Cref{tab:s-retr-all} extends \cref{tab:main}(a). The answer string, matched after lower-casing, collapsing whitespace
and removing thousands separators (every element for list answers), occurs on some gold or copy page for 112 of the 208
questions (53.8\%); for the others the answer-string hit cannot succeed. On the test sample alone, the all-question R@1
gap is $+3.2$ (BM25), $+1.3$ (E5), $+3.2$ (ColPali) and $+1.9$ (ColQwen2). Earlier versions scored a cross-page
subset of 48 questions in which, for two development questions, the evidence-aware hit credited a loose copy page,
contrary to the definition of $E(u)$; all retrieval tables now use the 46 questions with strict cross-page copies (R@1
gaps on the 48: $+22.9$, $+18.8$, $+16.7$, $+12.5$). The pre-set rule for the visual retrievers was stated on the 48;
it is met on both. The co-occurrence analysis (\cref{tab:s-cooc}) still uses the 48.

\begin{table*}[t]
  \centering\footnotesize\setlength{\tabcolsep}{3.5pt}
  \begin{tabular}{@{}llccccccccccc@{}}
    \toprule
    & & \multicolumn{5}{c}{All (208)} & \multicolumn{4}{c}{Cross-page copy (46)} & Test, cross-page (26) \\
    \cmidrule(lr){3-7}\cmidrule(lr){8-11}\cmidrule(l){12-12}
    Retriever & $k$ & Dataset & Gold & Answer & Evid. & Gap & Gold & Answer & Evid. & Gap {\scriptsize[95\% CI]} & Gap {\scriptsize[95\% CI]} \\
    \midrule
    BM25 & 1 & 32.2 & 37.0 & 36.1 & 42.3 & +5.3 & 19.6 & 34.8 & 43.5 & +23.9 {\scriptsize[13.0, 37.0]} & +19.2 {\scriptsize[3.8, 34.6]} \\
     & 5 & 63.0 & 69.7 & 53.8 & 72.1 & +2.4 & 52.2 & 54.3 & 63.0 & +10.9 {\scriptsize[2.2, 19.6]} & +19.2 {\scriptsize[7.7, 34.6]} \\
    E5 & 1 & 33.2 & 38.0 & 33.7 & 42.3 & +4.3 & 19.6 & 30.4 & 39.1 & +19.6 {\scriptsize[8.7, 30.4]} & +7.7 {\scriptsize[0.0, 19.2]} \\
     & 5 & 64.9 & 72.1 & 51.0 & 75.0 & +2.9 & 65.2 & 52.2 & 78.3 & +13.0 {\scriptsize[4.3, 23.9]} & +15.4 {\scriptsize[3.8, 30.8]} \\
    ColPali & 1 & 49.0 & 56.3 & 39.9 & 60.1 & +3.8 & 37.0 & 39.1 & 54.3 & +17.4 {\scriptsize[6.5, 30.4]} & +19.2 {\scriptsize[3.8, 34.6]} \\
     & 5 & 73.1 & 79.8 & 54.8 & 81.7 & +1.9 & 69.6 & 56.5 & 78.3 & +8.7 {\scriptsize[2.2, 17.4]} & +11.5 {\scriptsize[0.0, 23.1]} \\
    ColQwen2 & 1 & 50.0 & 60.1 & 40.9 & 63.0 & +2.9 & 43.5 & 39.1 & 56.5 & +13.0 {\scriptsize[4.3, 23.9]} & +11.5 {\scriptsize[0.0, 23.1]} \\
     & 5 & 78.4 & 84.1 & 56.3 & 85.6 & +1.4 & 73.9 & 56.5 & 80.4 & +6.5 {\scriptsize[0.0, 15.2]} & +7.7 {\scriptsize[0.0, 19.2]} \\
    \bottomrule
  \end{tabular}
  \caption{\textbf{Page retrieval with four retrievers and four hit definitions} (\MLB, recall, \%). Gap: evidence-aware
  minus verified gold.}
  \label{tab:s-retr-all}
\end{table*}

\paragraph{Citation judges.}
J1 asks, for every fact unit and candidate (5{,}584 pairs): ``Fact: $\langle$fact$\rangle$. Does this passage/image by
itself state this fact, with the same entity and the same value(s)? Answer yes only if it alone is enough to establish
the fact. Answer yes or no.'' Text candidates are given as text and image candidates as images. Its answers replace our
support relations; the benchmark's gold lists, the combinations and the contextual gold are unchanged. The pre-set
criterion required precision and recall each within 0.05 of our annotation against the human reference; neither judge
meets it (\cref{tab:s-judges}). The string-match relations use the numbers and quoted phrases of each unit's fact
sentence; 108 of 372 units contain none. An answer-level judge (J2: does the candidate provide evidence for the
reference answer?) used only for precision gives mean F1 0.334 and 0.488 under variant A and reverses 14.3\% and 13.0\%
of competitive pairs, against 0.552 for A with our relations.

\begin{table}[t]
  \centering\footnotesize\setlength{\tabcolsep}{2pt}
  \begin{tabular}{@{}lccccc@{}}
    \toprule
    & \multicolumn{3}{c}{Human reference (P / R)} & \multicolumn{2}{c}{vs.\ our annotation} \\
    \cmidrule(lr){2-4}\cmidrule(l){5-6}
    Relations from & All & Text & Image & P / R & $\kappa$ \\
    \midrule
    Our annotation & .74 / .87 & -- & -- & -- & -- \\
    Key-string match & .21 / .68 & -- & -- & .37 / .57 & -- \\
    Qwen2.5-VL-7B & .78 / .58 & .85 / .70 & .72 / .51 & .86 / .52 & .61 \\
    InternVL3-8B & .65 / .87 & .61 / .82 & .67 / .90 & .75 / .80 & .74 \\
    \bottomrule
  \end{tabular}
  \caption{\textbf{Support relations from string matching and VLM judges} (\MDR). Human reference: the 915 (quote, unit) relations
  of the 30 reviewed questions, positive when both reviewers judged them strict. The two judges agree at $\kappa=0.67$ (text 0.72, image
  0.61). --: not computed.}
  \label{tab:s-judges}
\end{table}

\paragraph{String masking.}
M\textsubscript{str}\textsuperscript{blk} masks the gold region and every text block of the six-page input that contains
a key string extracted in the first step of the pipeline (\cref{sec:method}). M\textsubscript{str}\textsuperscript{ctx}
instead adds random non-evidence blocks without key strings whose area matches what
M\textsubscript{str}\textsuperscript{blk} adds beyond the gold region ($\pm5\%$); it is run on copy items only, and
matching failed for 3 items and was not needed for 16 with no added area. The pre-set rule declared string masking
sufficient if it closes at least the share of the gap that M\textsubscript{auto} closes and if
M\textsubscript{str}\textsuperscript{ctx} lowers survival by less than 10 points; all three models meet both
(\cref{tab:s-string}). This shows that the string rule removes the answer as effectively, not that the masked set
isolates the fact: random blocks control for area, not for context that shares the key string. On copy items, M\textsubscript{str}\textsuperscript{blk} masks on average 3.1 times the area of
M\textsubscript{auto}, and 27.5\% (86/313) of its added blocks overlap an annotated or adjudicated copy, against 60.6\%
(57/94). The comparison asks whether the string rule suffices rather than removing only the verifier: both use the same
extracted keys, but M\textsubscript{auto}'s candidates also include blocks ranked by BM25 without a key string and are
capped in number. Within the key-string candidates of the original pipeline, verification raises block precision from
38.6\% (78/202) to 76.1\% (51/67). Controls carry no annotated copy, so their flips measure only the cost of over-masking.

\begin{table}[t]
  \centering\footnotesize\setlength{\tabcolsep}{1.5pt}
  \begin{tabular}{@{}lccc@{}}
    \toprule
    & Qwen2.5-VL & InternVL3 & Text model \\
    \midrule
    Survival, M\textsubscript{str}\textsuperscript{blk} & 20.7 & 19.2 & 8.0 \\
    Gap closed, M\textsubscript{str}\textsuperscript{blk} & 106 {\scriptsize[79, 144]} & 107 {\scriptsize[78, 150]} & 121 {\scriptsize[94, 167]} \\
    Gap closed, M\textsubscript{auto} & 78 {\scriptsize[50, 100]} & 79 {\scriptsize[50, 108]} & 79 {\scriptsize[46, 107]} \\
    M\textsubscript{gold} / M\textsubscript{str}\textsuperscript{ctx} survival & 90.9 / 95.5 & 84.2 / 89.5 & 76.2 / 85.7 \\
    $n$ (ctx) & 22 & 19 & 21 \\
    Control flips: str / auto & 5.9 / 2.0 & 11.1 / 5.6 & 11.4 / 2.3 \\
    \bottomrule
  \end{tabular}
  \caption{\textbf{Masking every key-string block without verification} (six-page input, \%). Gap closed: share of the
  M\textsubscript{gold}--M\textsubscript{all}\textsuperscript{blk} gap; values above 100 mean survival below
  M\textsubscript{all}\textsuperscript{blk}. Control flips: share of correctly answered controls whose correctness
  differs from that under M\textsubscript{gold}, in either direction. This is not the net accuracy loss B3 of
  \cref{tab:s-criteria}.}
  \label{tab:s-string}
\end{table}

\paragraph{Ablation-based attribution (ContextCite).}
We ran the official ContextCite implementation (commit c11f8ac; default Lasso, $\alpha=0.01$) with Qwen3-4B-Instruct-2507
on the six-page text inputs of the masking experiment. Sources are the text blocks used by
M\textsubscript{auto}; each ablation keeps every block independently with probability 0.5; the target is the model's own
answer under M0, scored for all 69 correctly answered items (25 copy, 44 control) with 64 and 128 ablations. One copy
item whose gold region is not text is excluded from gold statistics. The token log-probabilities were aggregated in
float64 because the official float32 aggregation overflows for highly confident answers; nothing else was changed. The
threshold, the median raw score of control gold blocks, was computed and frozen after the control runs and before
any copy item was run. Raw scores are not normalised for answer length, so the score contrast is descriptive; ranks are
scale-free. Block scores correlate only moderately between 64 and 128 ablations (Spearman 0.46, top-3 Jaccard 0.38), and
the default Lasso raised convergence warnings in some fits; \cref{tab:s-contextcite} should be read as aggregate
patterns.

\begin{table}[t]
  \centering\footnotesize\setlength{\tabcolsep}{2pt}
  \begin{tabular}{@{}lcc@{}}
    \toprule
    128 ablations & Copy (24) & Control (44) \\
    \midrule
    Gold block raw score & 10.1 {\scriptsize[6.2, 14.3]} & 32.8 {\scriptsize[25.3, 41.4]} \\
    All instances of the fact, summed & 30.6 {\scriptsize[23.2, 38.9]} & -- \\
    Best rank of a gold block & 13.0 {\scriptsize[3.0, 29.4]} & 2.7 {\scriptsize[1.3, 5.2]} \\
    A gold block in the top 3 & 70.8 {\scriptsize[50.0, 88.0]} & 88.6 {\scriptsize[78.6, 97.7]} \\
    Gold block below, set above threshold & 46.7 {\scriptsize[29.2, 63.8]} & -- \\
    \quad gold set also below & 25.0 {\scriptsize[8.7, 41.7]} & -- \\
    \bottomrule
  \end{tabular}
  \caption{\textbf{ContextCite under copies} (text model, correctly answered items; rates in \%). Threshold: median
  control gold-block score. ``Gold block below, set above threshold'' is the share of an item's gold blocks that meet
  the condition, computed within each item and averaged over the 24 items; it is not a share of items. --: not defined for controls, which have no copy. Intervals: bootstrap over questions.}
  \label{tab:s-contextcite}
\end{table}

\section{Released data}
\label{sec:s-release}
Code, annotations and frozen experimental outputs accompany this preprint as an ancillary archive (\texttt{code-and-data.7z}). The release includes two JSONL files and a small scoring library. \emph{\MLB}: 220 questions with split, document type,
applicability, dataset and verified gold pages, gold-page errors, and per unit the fact, key strings and every instance
(page, modality, role, copy form, strict or loose, boxes in PDF points where located). \emph{\MDR}: 200 questions with
gold quotes and, per unit, the strict, loose and combination quotes, plus contextual gold. The library computes
evidence-aware recall@$k$, the Quote F1 variants above, and the region list for set-level masking, and a script
reproduces the main paper's prevalence, re-scoring and retrieval numbers. Annotations follow the source benchmarks'
CC BY-NC 4.0 license; no document images or PDFs are redistributed.
Editable PowerPoint figure sources are provided in the ancillary archive \texttt{editable-figures.7z}.

\end{document}